\documentclass[final,3p,review,times]{elsarticle}

\usepackage[commandnameprefix=always,markup=default,authormarkup=none]{changes}
\usepackage{color} 
\definechangesauthor[name={Reviewer 1}, color=blue]{R1}

\usepackage{amssymb}
\usepackage{amsmath}
\usepackage{tabularx}
\usepackage{times}  
\usepackage{helvet}  
\usepackage{courier}  
\usepackage[hyphens]{url}  
\usepackage{graphicx} 
\usepackage{natbib}  
\usepackage{caption} 
\usepackage{subcaption}
\usepackage{booktabs}
\usepackage{diagbox}
\usepackage{color}
\usepackage{soul}
\usepackage{algorithm}
\usepackage{algorithmic}
\usepackage{newfloat}

\usepackage{listings}
\journal{Neural Networks}

\begin{document}

\begin{frontmatter}


\title{PVBF: A Framework for Mitigating Parameter Variation Imbalance \\
in Online Continual Learning}
\author[1]{Zelin Tao} 
\author[1,4]{Hao Deng\corref{cor1}}
\ead{denghao1984@tongji.edu.cn}
\author[2]{Mingqing Liu}
\author[3]{Lijun Zhang}
\author[1,4]{Shengjie Zhao}

\affiliation[1]{organization={School of Computer Science and Technology, Tongji University},
            city={Shanghai},
            postcode={201804}, 
            country={China}}

\affiliation[2]{organization={College of Electronic and Information Engineering, Tongji University},
            city={Shanghai},
            postcode={201804}, 
            country={China}}

\affiliation[3]{organization={National Key Laboratory for Novel Software Technology, Nanjing University},
            city={Nanjing},
            postcode={210023}, 
            country={China}}

\affiliation[4]{organization={Engineering Research Center of Key Software Technologies for Smart City Perception and Planning, Ministry of Education},
            city={Shanghai},
            postcode={201804}, 
            country={China}}

\cortext[cor1]{Corresponding author}

\begin{abstract}
Online continual learning (OCL), which enables AI systems to adaptively learn from non-stationary data streams, is commonly achieved using experience replay (ER)-based methods that retain knowledge by replaying stored past during training. However, these methods face challenges of prediction bias, stemming from deviations in parameter update directions during task transitions. This paper identifies parameter variation imbalance as a critical factor contributing to prediction bias in ER-based OCL. Specifically, using the proposed parameter variation evaluation method, we highlight two types of imbalance: correlation-induced imbalance, where certain parameters are disproportionately updated across tasks, and layer-wise imbalance, where output layer parameters update faster than those in preceding layers. To mitigate the above imbalances, we propose the Parameter Variation Balancing Framework (PVBF), which incorporates: 1) a novel method to compute parameter correlations with previous tasks based on parameter variations, 2) an encourage-and-consolidate (E\&C) method utilizing parameter correlations to perform gradient adjustments across all
parameters during training, 3) a dual-layer copy weights with reinit (D-CWR) strategy to slowly update output layer parameters for frequently occuring sample categories. Experiments on short and long task sequences demonstrate that PVBF significantly reduces prediction bias and improves OCL performance, achieving up to 47\% higher accuracy compared to existing ER-based methods.

\end{abstract}




\end{frontmatter}



\section{Introduction}

Learning constitutes the cornerstone of intelligent systems, enabling their adaptation to dynamic environments. Humans exemplify this adaptability through their ability to continuously acquire and integrate new knowledge while retaining prior experiences \cite{wang2024comprehensive}. In contrast, artificial intelligence systems, particularly those based on deep neural networks (DNNs), face a significant limitation known as catastrophic forgetting, where the acquisition of new information leads to the erosion of previously learned knowledge \cite{van2019three}, thereby hindering their capacity for sequential learning. To mitigate catastrophic forgetting and enable human-like learning capabilities, \textit{continual learning} (CL) methods have been extensively studied \cite{kirkpatrick2017overcoming,li2017learning,chaudhry2018riemannian}. Due to privacy concerns or resource constraints~\cite{mai2022online}, training data in more realistic online environments is presented as dynamic, one-pass streaming data \cite{soutif2023comprehensive, wang2024comprehensive}. Thus, CL models need to not only address catastrophic forgetting but also tackle the challenge of insufficient training, which can lead to prediction bias. Existing effective methods primarily employ strategies such as replaying past samples or representations, commonly referred to as experience replay (ER)-based methods \cite{rebuffi2017icarl,isele2018selective,chaudhry2019continual,pellegrini2020latent}. In recent years, research on ER-based methods has been considered crucial for enabling DNNs to continuously acquire new knowledge in dynamic and non-stationary sequential tasks. However, existing ER-based methods still suffer from significant prediction bias when applied to online continual learning (OCL) in sequential tasks. To address this issue, we propose an ER-based framework with bias-correction  strategies to improve the OCL performance of the models.
\begin{figure*}[h!]
    \includegraphics[width=1\textwidth,height=0.4\textwidth]{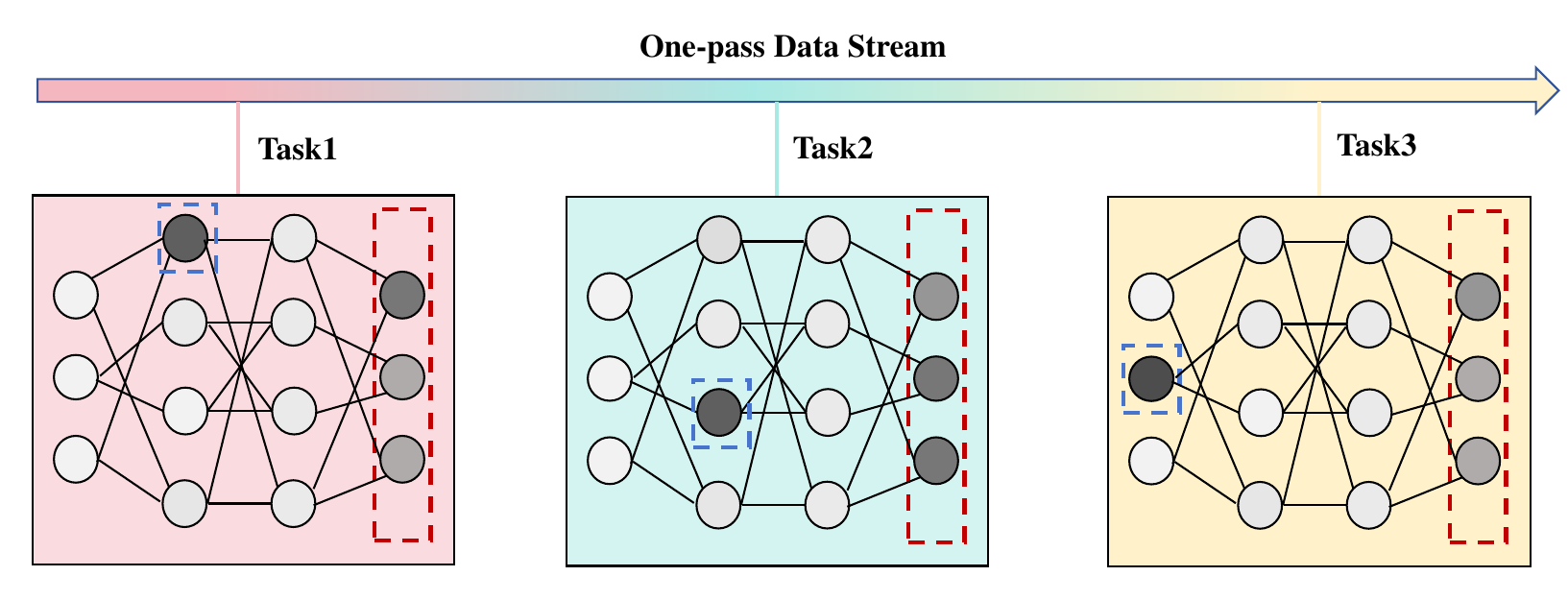}
    \centering
    \caption{Schematic diagram of parameter variation imbalance in OCL. An exemplary four-layer network is trained through a data stream with three tasks. The shading of the neurons indicates the magnitude of their variations during the training of the task, with darker colors representing greater variations. The blue dashed boxes present the  correlation-induced imbalance, where a small subset of parameters exhibits significantly larger variations during certain tasks. The red dashed boxes highlight the issue of layer-wise imbalance, where the average variation of the final linear layer in the network is significantly higher than that of adjacent layers.}
    \label{fig:intofig}
    \end{figure*}

Prediction bias is strongly linked to parameter variations during training \cite{sun2019optimization}. In OCL scenarios, the non-stationary nature of sequential tasks poses additional challenges, leading to more intricate and uneven parameter variations compared to conventional training settings \cite{Castro_2018_ECCV}. Existing ER-based methods primarily aim to mitigate task interference or preserve learned knowledge \cite{buzzega2020dark,caccia2021new}, but they lack explicit evaluation of parameter variations. As a result, these methods fail to effectively link parameter updates with task-specific requirements, limiting their ability to guide precise gradient adjustments and address prediction bias. In this work, as illustrated in Fig.~\ref{fig:intofig}, we investigate the issue of parameter variation imbalance in OCL and identify two key phenomena. First, during training on OCL tasks, subsets of parameters undergo disproportionately large variations due to their strong correlations with task-specific memorization. We term this the correlation-induced imbalance, where these parameters are improperly optimized for individual tasks, leading to uneven learning and interference across tasks. Second, the parameters of the output classifier exhibit significantly larger variations compared to adjacent layers, an issue we denote as the layer-wise imbalance. Our findings highlight that these imbalances, arising from uneven gradient updates across parameters, are critical contributors to prediction bias in OCL scenarios and underscore the need for systematic investigations into these dynamics.

For the correlation-induced imbalance issue, several CL approaches have investigated it, particularly from the perspective of parameter isolation. These methods suggest that different parameters of neural networks exhibit varying levels of correlation for memorizing previous tasks \cite{fernando2017pathnet,mallya2018packnet}. With CL parameter isolation methods, masks are typically used to freeze or adjust parameters closely related to previous learning tasks \cite{mallya2018piggyback,hurtado2021optimizing},  demonstrating significant effectiveness in improving overall model accuracy. However, these methods require allocating distinct sets of parameters for each task, leading to increased storage consumption as the number of tasks grows, which makes them less suitable for the dynamic, non-stationary sequence tasks targeted by OCL. In contrast, ER-based methods integrate new and old knowledge by mixing input samples to correct the overall parameter variation direction, effectively mitigating prediction bias while maintaining relatively low resource consumption. However, parameter updates in ER-based OCL methods may still deviate from the optimal direction during task switches due to the non-independent and identically distributed (IID) nature of the learning samples. One of the underlying causes is the insufficient understanding of the correlation between parameter variations and previous task memory. Existing work lacks in-depth investigation of identifying crucial parameters closely related to specific tasks and developing effective strategies to prevent these parameters from being improperly updated.

For the layer-wise imbalance issue, it stems from the observation that during the short task sequence OCL training process, parameters in the output classifier, i.e., the output layer parameters, tend to update more rapidly compared to other parameters \cite{chrysakis2023online}. This phenomenon can lead to a significant imbalance in parameter variations which further leads to prediction bias in the model \cite{wu2019large}. Existing research primarily focuses on improving the output layer individually based on experience replay \cite{pellegrini2020latent}, e.g., adding a bias correction layer after the output layer \cite{wu2019large}, using a surrogate classifier instead of the output layer during training \cite{chrysakis2023online}, and developing unique update strategies for the output layer \cite{lomonaco2017core50}. For instance, CWR* is a commonly adopted strategy for prediction bias correction by performing frequency-based magnitude adjustment in the output layer \cite{lomonaco2020rehearsal}. However, the first two methods inevitably increase the network complexity and the associated training costs, while the latter may introduce additional time expenditure and lead to suboptimal results. Designing an effective update strategy for the output layer that better adapts to dynamically non-stationary tasks remains a challenge. 

In this paper, we propose a Parameter Variation Balancing Framework (PVBF), which updates overall model parameters and output layer parameters at different stages to address the above two issues in OCL. To mitigate the correlation-induced imbalance, PVBF introduces Parameter Correlation Calculation (ParamCC) to quantify the correlation between each parameter and previous tasks. Building on this correlation measure, we propose Encourage and Consolidate (E\&C), a strategy that assigns adaptive gradient descent coefficients to parameters. This approach encourages the network to rapidly update parameters with low correlation to previous tasks while consolidating the memory of parameters with high correlation, thereby mitigating the prediction bias caused by the correlation-induced imbalance. To tackle layer-wise imbalance, we propose Dual-layer Copy Weights with Reinit (D-CWR), an improvement over the CWR* strategy. Inspired by human memory mechanisms (sensory memory, short-term memory, and long-term memory), D-CWR employs a two-stage consolidation process \cite{nairne2003sensory},  which can effectively mitigate the prediction bias caused by the output classifier in short task sequence OCL scenarios. 
The main contributions of this paper can be summarized as follows.

\begin{itemize}
\item[$\bullet$] It provides an evaluation method of parameter variations by capturing parameters' relative changes  during training. On this basis, we identify parameter variation imbalance in OCL through two aspects: correlation-induced imbalance and layer-wise imbalance, which provides prerequisites for optimized learning strategy design.

\item[$\bullet$] It proposes a two-phase approach to mitigate correlation-induced imbalance at the overall parameter level of neural networks. First, Parameter Correlation Calculate (ParamCC) quantifies the correlation between parameter variations and the memory of previously learned tasks. Then, Encourage and Consolidate (E\&C) strategy adjusts gradients by considering correlations as divisors to fine-tune the parameter updates. Together, parameter variation direction can be effectively corrected without huge network complexity increase.

\item[$\bullet$] It proposes a bio-inspired Dual-layer Copy Weights with Re-init (D-CWR) method to mitigate layer-wise imbalance. Drawing inspiration from the human memory mechanism, D-CWR employs memory consolidations at two layers, i.e., from sensory to short-term and from short-term to long-term memory, which further avoids forgetting due to rapid parameter updates in the output classifier over existing methods.

\item[$\bullet$] It presents a Parameter Variation Balancing Framework (PVBF) by integrating ParamCC, E\&C and D-CWR. Experiments conducted in both short and long task sequence OCL scenarios show that PVBF achieves an average accuracy improvement of 31\%-47\% over the ER method. Especially, it reaches 97.5\% of the IID method's accuracy using only 500 replay samples on the MiniImageNet dataset. PVBF also exhibits general outperformance in offline CL scenarios among classical methods.
\end{itemize}

\section{Related Work}

In this section, we briefly review three key categories of research relevant to the current work: traditional CL methods, ER-based methods, and output correction methods. The strengths and weaknesses of each method are briefly discussed.
\subsection{CL Methods}

Contemporary CL methodologies predominantly fall into three categories \cite{van2019three}. Regularization-based approaches penalize parameter updates to enforce convergence within a shared representation space across diverse tasks \cite{kirkpatrick2017overcoming,chaudhry2018riemannian,zenke2017continual}. These approaches tackle the problem of catastrophic forgetting by constraining the parameter update methods. However, they faces challenges in balancing between stability and plasticity, often resulting in suboptimal performance and high computational cost. Memory-based strategies integrate prior task knowledge through sample or representation-based memory adjustments during training \cite{lopez2017gradient,chaudhry2018efficient,shin2017continual,riemer2018learning,wang2022foster,bonicelli2022effectiveness}. While these methods effectively retain valuable knowledge across tasks, they require storing a subset of training samples and may introduce prediction bias to some extent due to issues with the training data distribution. Dynamic structures-based approaches adapt network architectures to ensure task-specific parameter isolation, accommodating the integration of novel tasks \cite{li2017learning,yoon2017lifelong,von2019continual,lomonaco2020rehearsal}.  These approaches address the issue of interference by isolating task-specific knowledge, but they struggle with computational efficiency and spatial scalability as the number of tasks increases. Thus, methods in OCL often combine with memory-based strategies. However, challenges remain in striking a balance between memory usage, computational efficiency, and maintaining performance on non-stationary data streams.

\subsection{ER-based Methods}

Existing effective approaches in OCL typically adopt strategies that involve replaying samples or representations. To mitigate catastrophic forgetting caused by task changes, ICaRL \cite{rebuffi2017icarl} combines distillation loss with binary cross-entropy, classifying samples based on nearest-class prototypes computed from buffered data representations, which is suitable for class incremental learning scenarios where each task is sufficiently trained. However, in OCL scenarios, iCaRL often underperforms due to insufficient training on newer tasks. ER \cite{chaudhry2019continual} employs a fixed-size replay buffer, randomly replaying a subset of samples. Despite its simplicity, ER faces challenges in maintaining performance when learning from both replayed samples and data stream samples simultaneously. To address this issue, GDumb \cite{prabhu2020gdumb} maintains a class-balanced memory pool and trains the model exclusively on these samples, although the size of the memory pool often constrains its effectiveness. MIR \cite{aljundi2019online} introduces an alternative improvement to ER by selecting samples that maximize the increase in loss during replay. This method further reduces prediction bias, albeit at the cost of increased computational burden. For loss calculation, DER++ \cite{buzzega2020dark} employs distillation loss on logits to enforce consistency over time, while ACE \cite{caccia2021new} mitigates sudden representation changes using an asymmetric update rule. Although DER++ and ACE are promising in stabilizing learned knowledge, they fail to resolve the issue of spatially or structurally imbalanced parameter updates across the network, which can lead to significant prediction bias. 

\subsection{Output Correction Methods}
In OCL scenarios, prediction bias is closely related to rapid updates of the output classifier during backpropagation \cite{wu2019large}. OBC \cite{chrysakis2023online} independently optimizes output classifier to correct significant prediction bias during training. AR1* \cite{lomonaco2020rehearsal} combines latent replay methods with optimized output classifier updates from CWR* to enhance performance. The CWR* method aligns the parameters in the output classifier with individual categories. It adjusts the update magnitude of these parameters based on the ratio of the frequency of data from a particular class in past occurrences to its frequency in a single training iteration, thereby correcting prediction bias. These strategies address the bias in the output classifier but do not simultaneously correct the bias in other parameters, leaving room for further improvement. 

Although the three types of methods discussed above do not directly address the issue of parameter variation imbalance, they provide foundational ideas that have informed our approach to solving this problem. ER effectively mitigates catastrophic forgetting caused by non-IID data distributions in streaming tasks, while asymmetric cross entropy (ACE) helps achieve more accurate and balanced gradient updates. These strategies contribute to balancing parameter variations during OCL training to some extent, and thus, we integrate these concepts into our proposed framework. Furthermore, previous work on special handling of the output classifier has inspired our approach to addressing the issue of layer-wise imbalance.
\section{Methodology}

In this section, a detailed explanation of parameter variation imbalance is first presented, including its two specific forms and their manifestations in the OCL scenario. The proposed PVBF is then introduced, along with its applicable learning settings. Finally, detailed design and implementations of innovative methods incorporated into PVBF, namely ParamCC, E\&C, and D-CWR, are described, respectively.
\subsection{Parameter Variation Evaluate}
In OCL, model parameters continuously evolve with the training of various tasks, and the ultimate result of this variation determines the model's adaptability across the entire online data stream. Therefore, we aim to calculate the variations of its parameters upon the completion of training for each task in the OCL context, revealing the patterns of parameter variation.

To record the parameters optimized through task $k$ (where $k
\in \{1,\cdots,K\}$ and $K$ denotes the total number of tasks), we capture the model state at the moment of the first occurrence of $(k+1)-{\rm th}$ task denoted as $\theta_{1.k}, \ldots, \theta_{M,k}$ (where $M$ represents the total number of parameters in the neural network). To model the variation of parameters, we first define the variation in parameter ranked $m$ ($m\in \{1,\cdots,M\}$) between task \( k \) and \( k-1 \) as $\delta_{m,k}$, which is calculated using the Manhattan distance as

\begin{equation}
\delta_{m,k} = |\theta_{m,k} - \theta_{m,k-1}|,
\end{equation}
where $|\cdot|$ denotes the Manhattan distance operator.

Due to significant differences in gradient updates between different parts of the network, $\delta_{m,k}$ can vary considerably for different $m$. Consequently, relying solely on its numerical variation may not accurately capture the correlation between parameters and tasks. To address this issue, we first standardize $\delta_{m,k}$ for each parameter while preserving relative changes, thereby mitigating the impact of numerical imbalance:
\begin{equation}
\delta_{m,k}^{'} = \mathcal{S}(\delta_{m,k}),
\end{equation}
where $\mathcal{S}$ is a customized standardization function, and $\delta'_{m,k}$ is defined as the relative change of parameter $m$ on task $k$.


Here we introduce three exemplary standardization functions utilized in identifying parameter variance imbalance. The first is the relative ratio (RR) function denoted by $\mathit{RR}(\cdot)$, which is expressed as
\begin{equation}
\Bar{\delta}_{k} = \frac{\sum_{m=1}^M \delta_{m,k}}{M} , \quad 
\mathit{RR}(\delta_{m,k}) = \frac{\delta_{m,k}}{\Bar{\delta}_{k}},
\end{equation}
where $\Bar{\delta}_{k}$ represents the mean parameter variation for task $k$. This standardization enables $\delta'_{m,k}$ to capture relative parameter variation patterns without being affected by the absolute values of gradient updates. Additionally, by using a proportional form, this method mitigates the impact of different network architectures while intuitively highlighting imbalances in parameter variations, making it easier to identify parameters with disproportionately large or irregular updates.

As an alternative to the RR approach, Z-score (ZS) standardization method focuses on scaling the parameter variations with respect to their mean and standard deviation. The ZS standardization denoted by $\mathit{ZS}(\cdot)$ is defined as
\begin{equation}
\sigma_k = \sqrt{\frac{1}{M}\sum_{m=1}^M(\delta_{m,k}-\Bar{\delta}_{k})^2}, \quad \mathit{ZS}(\delta_{m,k}) = \frac{\delta_{m,k}-\Bar{\delta}_{k}}{\sigma_k},
\end{equation}
where $\sigma_k$ represents the standard deviation of the parameter variations for task $k$. ZS method accounts for both the central tendency and the spread of variations. Unlike the RR approach, the ZS method does not rely on task-specific variation proportions but instead standardizes the variations based on statistical properties, making it particularly useful for highly heterogeneous variations or those with significant outliers.

The last standardization method is robust scaler (RS), which is also particularly advantageous for handling significant outliers or skewed distributions of parameter variations. The RS method standardizes the variations based on the median and interquartile range (IQR), which is defined as
\begin{equation}
\mathit{RS}(\delta_{m,k}) = \frac{\delta_{m,k}-Median(\delta_{k})}{IQR(\delta_{k})},
\end{equation}
where \({Median}(\delta_k)\) represents the median and \({IQR}(\delta_k)\) is the interquartile range (the difference between the 75th and 25th percentiles) of parameter variations for task \(k\). The RS method is highly resilient to outliers, as the median and IQR are less sensitive to extreme values than the mean, ensuring robust parameter scaling in noisy or heavy-tailed distributions.

\begin{figure*}[h!]
    \centering
    \begin{subfigure}{0.24\textwidth}
        \centering
        \includegraphics[width=\linewidth]{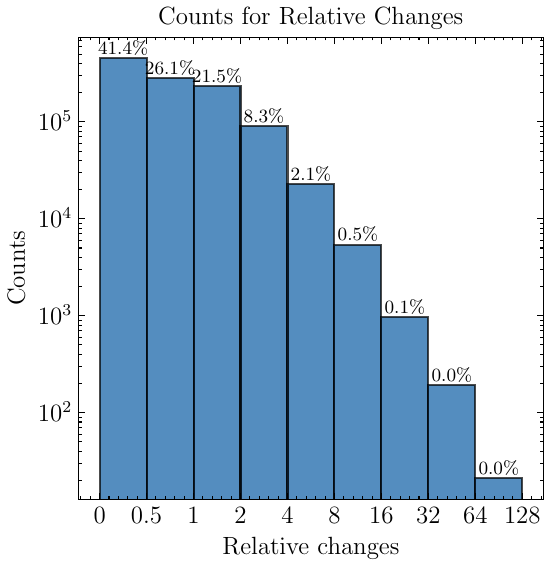}
        \caption{Task 1}
        \label{fig:sub1}
    \end{subfigure}
    \begin{subfigure}{0.24\textwidth}
        \centering
        \includegraphics[width=\linewidth]{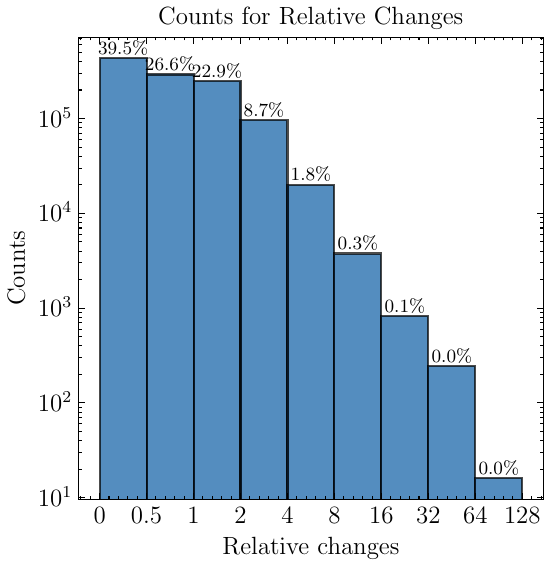}
        \caption{Task 2}
        \label{fig:sub2}
    \end{subfigure}
    \begin{subfigure}{0.24\textwidth}
        \centering
        \includegraphics[width=\linewidth]{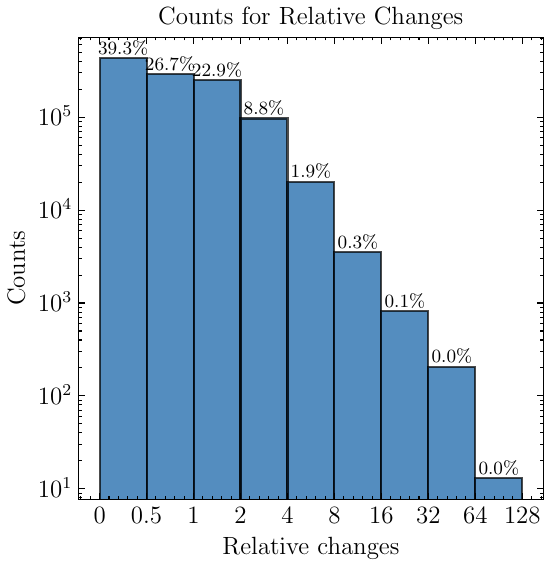}
        \caption{Task 3}
        \label{fig:sub3}
    \end{subfigure}
    \begin{subfigure}{0.24\textwidth}
        \centering
        \includegraphics[width=\linewidth]{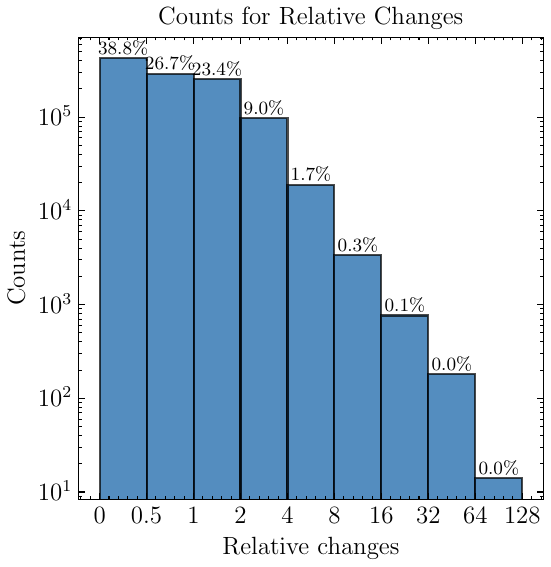}
        \caption{Task 4}
        \label{fig:sub4}
    \end{subfigure}
    \caption{Neuron counts for different relative changes $\delta_{m,k}^{'}$}
    \label{fig:four_images}
\end{figure*}

\begin{figure*}[h!]
    \centering
    \begin{subfigure}{0.24\textwidth}
        \centering
        \includegraphics[width=\linewidth]{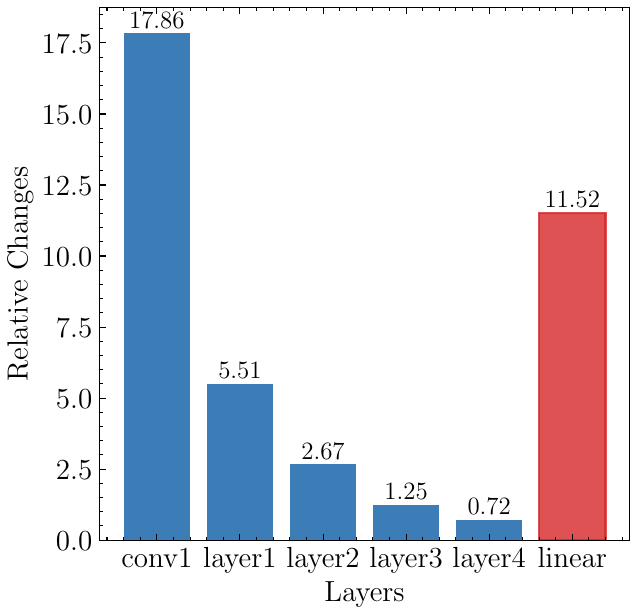}
        \caption{Task 1}
        \label{fig:sub1}
    \end{subfigure}
    \begin{subfigure}{0.24\textwidth}
        \centering
        \includegraphics[width=\linewidth]{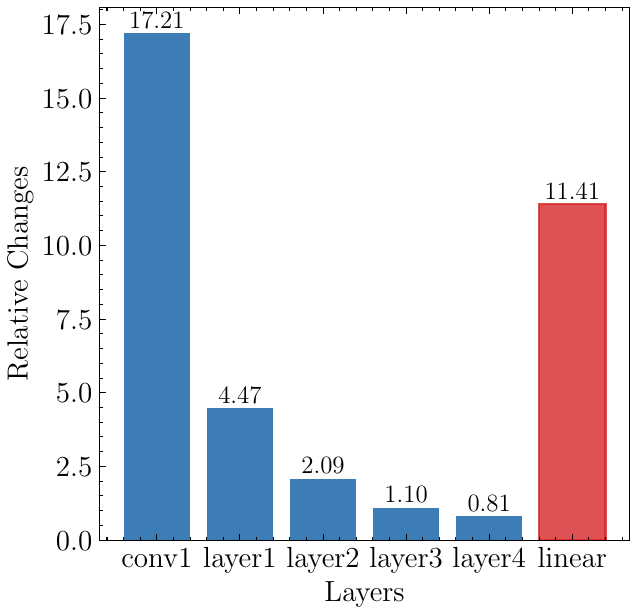}
        \caption{Task 2}
        \label{fig:sub2}
    \end{subfigure}
    \begin{subfigure}{0.24\textwidth}
        \centering
        \includegraphics[width=\linewidth]{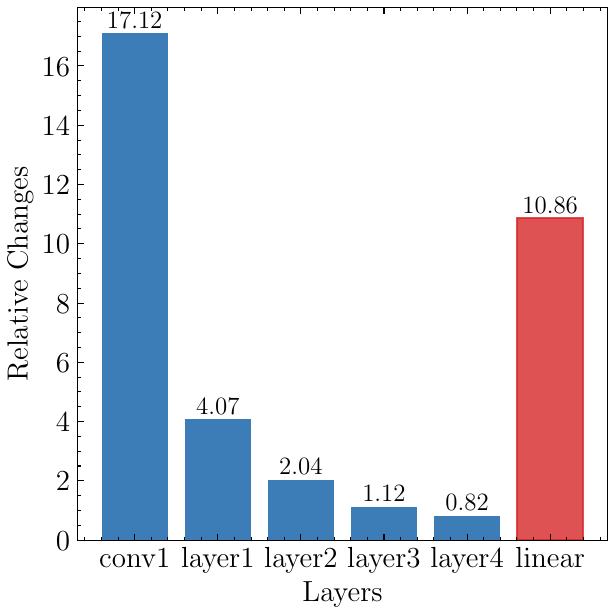}
        \caption{Task 3}
        \label{fig:sub3}
    \end{subfigure}
    \begin{subfigure}{0.24\textwidth}
        \centering
        \includegraphics[width=\linewidth]{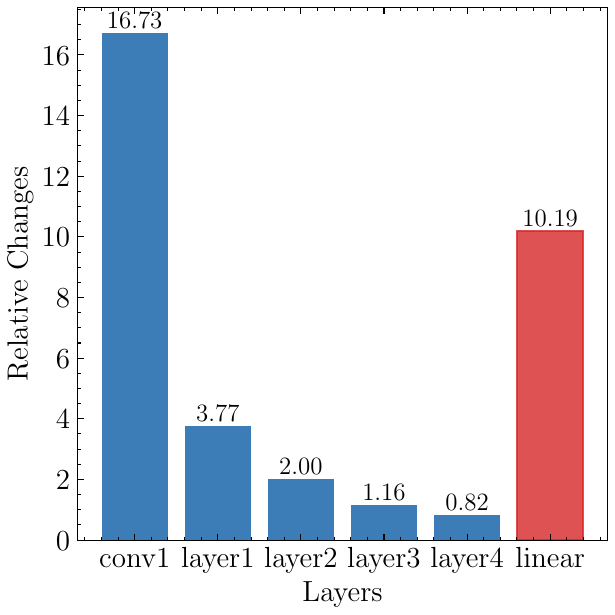}
        \caption{Task 4}
        \label{fig:sub4}
    \end{subfigure}
    \caption{Average relative changes in parameters across different layers}
    \label{fig:four_images2}
\end{figure*}

Using the above standardization methods for obtaining $\delta_{m,k}^{'}$, we recorded the relative changes at the end of $1\sim 4$ task in a sequence of short tasks (a total of 5 tasks) during continual training on Cifar10 \cite{krizhevsky2009learning} with ER method \cite{chaudhry2019continual} using a backbone of reduced-Resnet18 \cite{aljundi2019online}. Through numerical analysis, RR intuitively demonstrates the relative differences in parameter updates, highlighting potential imbalances in both the overall and hierarchical patterns of parameter variations, while maintaining minimal computational complexity. Therefore, for all the analysis and experiments presented below, we specifically used the RR standardization method. 
The results obtained using the other two standardization methods are provided in Appendix A. More importantly, our observations reveal two distinct imbalances within the OCL scenario, as shown in Figs. \ref{fig:four_images} and \ref{fig:four_images2}.

\textbf{Correlation-induced Imbalance}. Fig.~\ref{fig:four_images} illustrates the uneven distribution of parameter updates during OCL training, where a small fraction of parameters undergo significant changes, with some updates exceeding 64 times the mean, while over 65\% of parameters exhibit updates smaller than the mean ($\delta^{'}_{m,k}<1$). This indicates that most parameters remain inactive, contributing minimally to task learning. In contrast, a smaller subset of parameters which experience substantial updates correlates closely with task memory. Such imbalances suggest inefficient parameter utilization, where inactive parameters are underutilized, and heavily updated parameters are at risk of causing catastrophic forgetting. Therefore, subsequent training should focus on better engaging inactive parameters while safeguarding those critical for task retention.

\textbf{Layer-wise Imbalance}. Fig.~\ref{fig:four_images2} demonstrates that in CL, the output classifier, i.e. the last linear layer, exhibits consistently higher parameter update rates compared to earlier feature extraction layers. This occurs because backpropagation begins at the classifier, which directly addresses prediction bias, particularly when input samples are imbalanced. Consequently, the classifier undergoes more frequent and larger updates, amplifying prediction biases and potentially destabilizing learning. To mitigate this, balancing updates across layers or restricting output classifier updates can reduce this disparity and improve the overall stability of the model.

\subsection{Overview of Parameter Variation Balancing Framework}
\begin{figure*}[t!]
    \includegraphics[width=1\textwidth,height=0.55\textwidth]{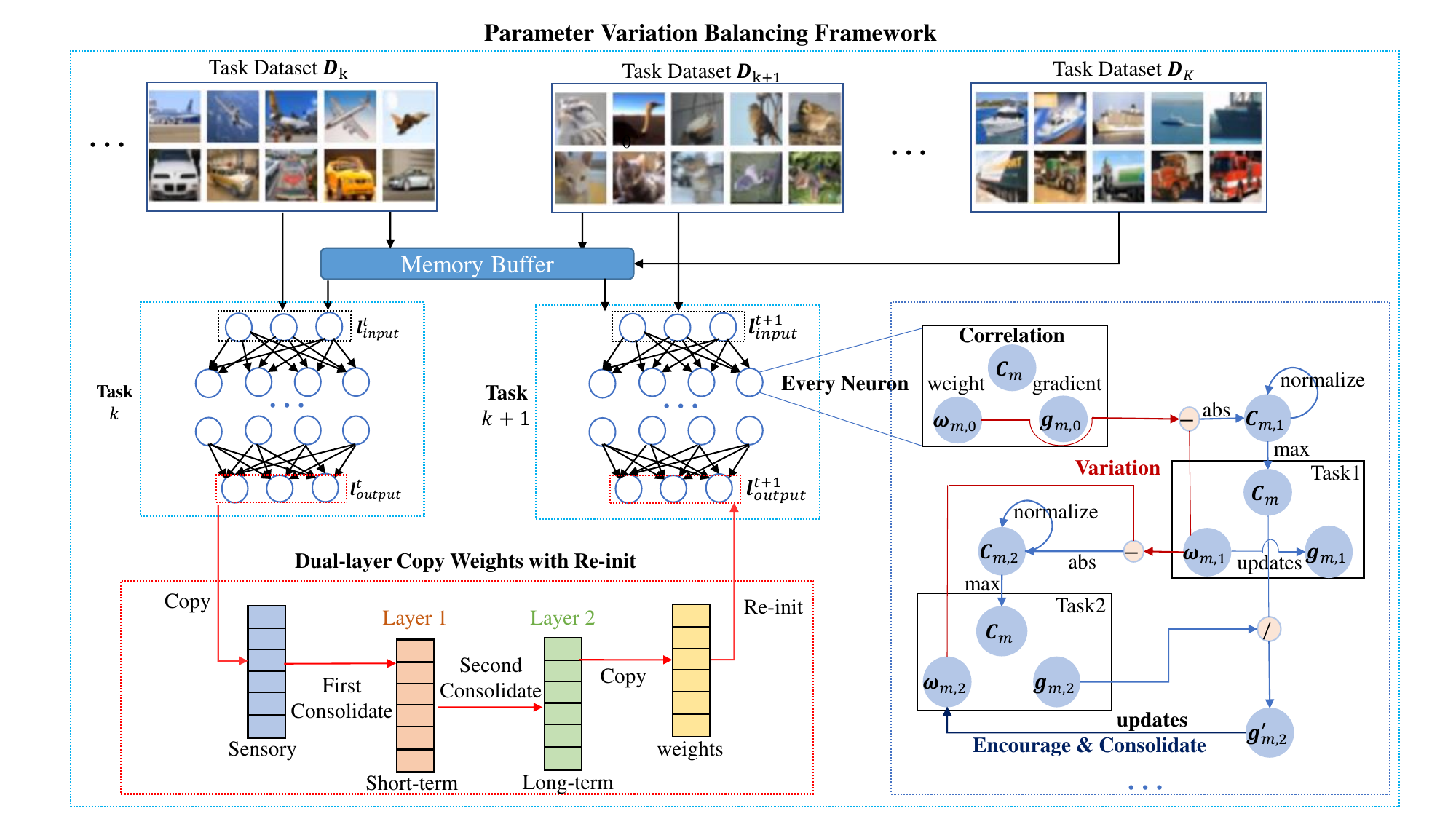}
    \centering
    \caption{Overall illustration of PVBF. PVBF is built upon a conventional experience replay framework and incorporates two main strategies for balancing parameter variations. First, the Encourage and Consolidate (E\&C) strategy dynamically adjusts all neurons in the network during training. This strategy leverages the parameter correlations obtained through the Parameter Correlation Calculate (ParamCC) method during training to adjust the gradients computed by the Stochastic Gradient Decent (SGD) process. Second, the Dual-layer Copy Weights with Re-init (D-CWR) strategy is specifically applied to the output classifier, which progressively reinforces knowledge of individual class categories.}
    \label{f:overview}
    \end{figure*}

To address the two parameter variation imbalance issues outlined above, we propose PVBF designed for OCL scenarios. As in Fig.~\ref{f:overview}, PVBF primarily consists of two balancing strategies, E\&C and D-CWR\, to mitigate imbalanced parameter variations in both the overall and the output layer level during training, thereby enhancing the network's adaptability to non-stationary data streams. PVBF builds upon an ER framework, which retains a memory buffer storing part samples from previous tasks to aid in knowledge retention. For each neuron in each task, ParamCC is proposed to calculate correlation between parameter and previous tasks by capturing the variations in parameter weights during the training process. First, it employs a Manhattan distance-based metric to monitor relative parameter changes after each OCL task. Then, these changes are normalized using Min-Max normalization before and after task transitions to ensure a consistent correlation measurement \cite{jiawei2006data}. To reduce time and space overhead, only the maximum correlation value of each parameter is retained throughout training. Based on these correlation values, E\&C is then applied to update the gradients, encouraging low-correlation parameters to acquire new knowledge while consolidating high-correlation parameters to preserve previously learned knowledge. At the end of each task's training, the D-CWR strategy is applied to the output classifier. Parameters obtained through gradient descent are treated as sensory memory. Class-specific knowledge is then selectively transferred to a short-term memory pool through the first consolidation, followed by further consolidation into a long-term memory pool with a defined probability. Finally, the knowledge stored in long-term memory is used for prediction. This strategy slows the update rate of output layer parameters to reduce the prediction bias caused by rapid updates in this layer. 

Besides, our study focuses on CL in dynamic and non-stationary sequential tasks. Firstly, we adopt the CL setting, where a model with parameters $\boldsymbol{\Theta}$ must generalize well to test data without full access to previous training samples \cite{caccia2021new}. Each task $k$ consists of training data $
\mathbf{D}_{k} = \{\mathbf{X}_k,\mathbf{Y}_k\}$, where $\mathbf{X}_k$ presents the input data and $\mathbf{Y}_k$ the corresponding labels. To ensure task diversity, we follow the classic disjoint label setting, where label sets of different tasks do not overlap ($\mathbf{Y}_i \cap \mathbf{Y}_j = \emptyset$ for $i \not= j$). Building upon this, we further explore OCL setting, where training process is divided into several time steps, with each time step corresponding to a single batch of training data. At each time step $t$, the model receives a new batch of training data ($\mathbf{X}^{in}_{t},\mathbf{Y}^{in}_{t}$). Over time, the data distribution shifts as new tasks emerge. Specifically, at certain time steps $\{t_\alpha,t_\beta,...\} \in T$, the data transition from one task to another, which means that the continuous two sets of samples that the learner receives may belong to different tasks, namely $\{\mathbf{X}^{in}_{t_{\alpha-1}},\mathbf{Y}^{in}_{t_{\alpha-1}}\} \in D_k$ and $\{\mathbf{X}^{in}_{t_{\alpha}},\mathbf{Y}^{in}_{t_{\alpha}}\} \in D_{k+1}$. The training tasks occur sequentially, and task identifiers are not provided during evaluation. In this non-stationary setting, maintaining the neural network's classification performance across tasks is particularly challenging. We posit that preserving more task-specific network parameters' memory during learning while mitigating prediction bias from parameter updates during new task learning is critical to overcoming this challenge.


\begin{algorithm}[htbp]
\caption{Parameter Variation Balancing Framework (PVBF)}\label{alg:alg1}
\begin{algorithmic}
\STATE \textbf{Input} the hyperparameter $\alpha$, $\beta$, Learning rate $l$.
\STATE \textbf{Initialize} network parameters $ \boldsymbol{\Theta} $ ,gradients $\boldsymbol{g}$, and memory buffer size $\rm{M}$
\STATE \textbf{Initialize} $ \boldsymbol{\mathit{C}} $, $ \boldsymbol{M^s} $, $ \boldsymbol{M^c} $, $ \boldsymbol{M^l} $ and $ \boldsymbol{P} $ to zeroes
\STATE \textbf{for} each task $ k \in \{1,...,K\} $ \textbf{do}
\STATE \hspace{0.5cm} store network parameters $\boldsymbol{\Theta}$ in $\boldsymbol{\theta}_{k-1}$
\STATE \hspace{0.5cm} \textbf{for} each training batch at timestep $t$ $ (\mathbf{B}_t) $ \textbf{do}
\STATE \hspace{0.5cm} \hspace{0.5cm} receive $\mathbf{X}^{in}_t \sim \mathbf{B}_t$ from input stream, $\mathbf{X}^{re}_t \sim \mathbf{M}$ from memory buffer
\STATE \hspace{0.5cm} \hspace{0.5cm} Calculate $\mathcal{L}(\mathbf{X}^{bf}_t \cup \mathbf{X}^{in}_t)$
\STATE \hspace{0.5cm} \hspace{0.5cm} $\boldsymbol{g} \leftarrow SGD(\nabla \mathcal{L}, \boldsymbol{\Theta})$ //calculate gradients
\STATE \hspace{0.5cm} \hspace{0.5cm} \textbf{if $k > 1$} $\boldsymbol{g'} \leftarrow \boldsymbol{g} ,\boldsymbol{\mathit{C}} $\hfill $\rhd$ section 3.4  
\STATE \hspace{0.5cm} \hspace{0.5cm} $\boldsymbol{\boldsymbol{\Theta}} \leftarrow \boldsymbol{g'}, l$ 
\STATE \hspace{0.5cm} \textbf{end for}
\STATE \hspace{0.5cm} $\boldsymbol{\delta'} \leftarrow \boldsymbol{\Theta}, \boldsymbol{\theta}_{k-1}$ \hfill $\rhd$ section 3.2
\STATE \hspace{0.5cm} $\boldsymbol{C} \leftarrow \boldsymbol{\delta'}, \alpha,\beta$ \hfill $\rhd$ section 3.3
\STATE \hspace{0.5cm} update $ \boldsymbol{M^s} $, $ \boldsymbol{M^c} $, $ \boldsymbol{M^l} $ and $ \boldsymbol{P} $ using D-CWR  \hfill $\rhd$ section 3.5
\STATE \textbf{end for}
\end{algorithmic}
\label{alg1}
\end{algorithm}
We adopt ACE to compute the loss, which separately handles data from the data stream and the memory buffer~\cite{caccia2021new}. The total loss is given by
\begin{equation}
\mathcal{L}(\mathbf{X}^{bf}_t \cup \mathbf{X}^{in}_t) = \mathcal{L}^{re}(\mathbf{X}^{bf}_t, \mathbf{Y}^{old}_t \cup \mathbf{Y}^{curr}_t) + \mathcal{L}^{in}(\mathbf{X}^{in}_t, \mathbf{Y}^{curr}_t),
\end{equation}
where $\mathcal{L}$ denotes the total loss function, $\mathcal{L}^{in}$ and $\mathcal{L}^{re}$ represent the cross-entropy loss of the data stream and data obtained from the memory buffer, \( \mathbf{X}^{bf}_t \) and \( \mathbf{X}^{in}_t \) represent the subsets of training samples at time step \( t \), \( \mathbf{Y}^{old}_t \) denotes the classes encountered up to that point, and \( \mathbf{Y}^{curr}_t \) refers to the current training classes. Finally, a complete workflow of PVBF is provided in Algorithm 1. In the following, we will detail key designs of ParamCC, E\&C strategy, and D-CWR strategy, respectively.

\subsection{Parameter Correlation Calculate}
To reduce the impact of gradient updates on absolute parameter variations and clarify the relationship between parameter variations and their correlation with previous tasks, Min-Max normalization can be used to standardize parameter correlations within a fixed range. Based on this idea, we introduce ParamCC, a method for evaluating the association between specific neural network parameters and previously learned tasks. The correlation of each parameter $\theta_m$ to the memory of task $k$ can be expressed as follows:
\begin{equation}
C_{m,k} = \frac{\delta^{'}_{m,k} - \min(\boldsymbol{V}_k)}{\max(\boldsymbol{V}_k) - \min(\boldsymbol{V}_k)} \cdot (\beta - \alpha) + \alpha,
\end{equation}
where $C_{m,k}$ denotes the correlation of parameter $m$ with the memory of task $k$, and $\boldsymbol{V}_k = \{\delta_{1,k}^{'}, \ldots, \delta^{'}_{M,k}\}$ represents the set of relative changes. The hyperparameters $\alpha$ and $\beta$ define the range of correlation for the parameters with the tasks, with $C_{m,k}$ being normalized to the interval $[\alpha, \beta]$. 

By applying the aforementioned methods, we can derive a correlation measure $C_{m,k}$ for each parameter in the network with each previously trained task. To efficiently utilize this correlation measure and reduce computational and spatial overhead, we maintain only the maximum correlation of each parameter with the previous tasks throughout the training process as
\begin{equation}
C_m = \max_{r \in \{1, \ldots, k-1\}} C_{m,r},
\end{equation}
where $C_m$ represents the correlation of parameter $m$ with all previously experienced tasks, and its value can be computed in real time during the training process. The variable $k$ denotes the task currently being trained. 
\subsection{Encourage and Consolidate}

To mitigate catastrophic forgetting, we further integrate parameter correlation into the training process and propose a gradient correction method called E\&C. The core idea is to encourage rapid updates for parameters with low correlation to previous tasks while consolidating those with high correlation by decelerating their update rates. For time step $t$, the gradient $g_{m,t}$ obtained by neuron $m$ through gradient descent undergoes the following correction process:

\begin{equation}
g_{m,t}^{'} = \frac{g_{m,t}}{C_m},
\end{equation}
where $g'_{m,t}$ represents the adjusted gradient for neuron $m$ at time step $t$, taking into account the correlation of the parameters to previously learned tasks. This correction enables rapid updates for parameters with low correlation to previous tasks, thereby encouraging them to acquire new knowledge. Conversely, it ensures that parameters with high correlation to previous tasks are updated more cautiously, thereby consolidating their memory of those tasks. During the training process of online continual learning, we utilize the corrected parameters $ \{g'_{1,t}, \ldots, g'_{M,t}\} $~for parameter updates. This E\&C approach facilitates a balance between learning new tasks and retaining knowledge of old tasks, ultimately enhancing the model's adaptability and mitigating catastrophic forgetting.

\subsection{Dual-Layer Copy Weights with Re-init}
To minimize the phenomenon of parameter variation imbalance, in addition to the E\&C method designed for overall model-level corrections, PVBF incorporates a specialized correction method specifically targeting the output classifier. Inspired by the interactions among sensory memory, short-term memory, and long-term memory (denoted as $s$,$c$,$l$)~in the complementary learning systems (CLS) framework \cite{mcclelland1995there}, we propose an improved approach that enhances memory consolidation and reduces forgetting. Specifically, we introduce a D-CWR method that simulates the memory process after copying weights ,which involves two consolidation layers: the first layer simulates the hippocampal process of converting sensory memory into short-term memory, while the second layer models the neocortical process of transferring short-term memory into long-term memory. In the following, we first define the concept of sensory memory in the context of the output classifier, followed by an explanation of the short-term and long-term consolidation processes.

We consider that the sensory memory of the output classifier originates from the knowledge acquired during the training process, particularly the model parameter weights obtained through gradient descent. In experiments conducted in \cite{maltoni2019continuous}, a technique called mean-shift effectively normalizes the output classification layer model parameters, maintaining them within a certain range while emphasizing the features of individual classes ,which inspires our design of sensory memory. We define the sensory memory of the output classifier as the parameter weights assigned to each class after each training iteration, subtracted by the mean of the weights across all classes that appear in that iteration, specifically as follows:
\begin{equation}
    M^s_{j,t} = \omega_{j,t} - \frac{\sum_{k \in S_t} \omega_{k,t}}{|S_t|},
\end{equation}
where $M^s_{j,t}$ represents the sensory memory of the model for class $j$ , $ \omega_{j,t} $ denotes the $j$-th parameter of the output classifier, and $S_t$ represents the set of sample labels input to the model, at time step $t$\chdeleted{. O}\chadded{, o}perator $|\cdot|$ indicates calculating the cardinality of the set. Note that the mapping of the output classifier parameters to each class is determined by the properties of the softmax activation function. The softmax function connecting the final linear layer transforms the output layer parameters into classification probabilities for each class.

In CLS, the hippocampus is responsible for converting sensory memory into short-term memory by selectively processing certain information. The first layer of the D-CWR consolidation strategy simulates this process of the hippocampus. When sensory memory is received, D-CWR randomly selects one of the following two actions to execute: i) Integrating the current sensory memory with the existing short-term memory; ii) Retaining the previous short-term memory while withholding processing the current sensory memory. The model handles the short-term memory $M^c_{j,t}$ for class $j$ at time step $t$, which can be calculated as
\begin{equation}
M^c_{j,t} = \begin{cases} 
M^c_{j,t-1},  & \text{if }\epsilon \ge p \\
\frac{M^c_{j,t-1} \cdot \eta^c_{j,t}+ M^s_{j,t}}{\eta^c_{j,t}+1}, & \text{if }\epsilon <p
\end{cases},
\end{equation}
where $\epsilon$ is a random variable uniformly distributed in the interval $[0,1]$, $p$ is a hyperparameter representing the probability that the model converts sensory memory into short-term memory, and $\eta^c_{j,t}$ denotes the short-term memory retention coefficient for class $j$ at time step $t$.

The neocortex plays a key role in converting short-term memory into long-term memory, with more frequent review of a particular class leading to stronger memory consolidation. The second layer of the D-CWR consolidation strategy simulates this process of the neocortex by integrating short-term memory with existing long-term memory according to a specified retention ratio. The long-term memory $M^l_{j,t}$ for class $j$ at time step $t$ is expressed as
\begin{equation}
    M^l_{j,t} = \frac{M^l_{j,t-1} \cdot \eta^l_{j,t}+M^c_{j,t}}{\eta^l_{j,t}+1},
\end{equation}
where $\eta^l_{j,t}$ denotes the long-term memory retention coefficient for class $j$ at time step $t$, $\eta^l_{j,t}$ can be calculated alongside the short-term memory retention coefficient $\eta^c_j$ for class $j$ at time step $t$ as
\begin{equation}
    \eta^c_{j,t} = \frac{P_{j,t}}{U_{j,t}} ; \eta^l_{j,t} = \sqrt{\frac{P_{j,t}}{U_{j,t}}},
\end{equation}
with $P_{j,t}$ denotes the number of times class $j$ has appeared during the model training process at time step $t$, and $U_{j,t}$ represents the number of occurrences of class $j$ in the label set $S_t$.

To implement the D-CWR strategy, we present the algorithm that integrates the two consolidation stages, which begins by initializing memory vectors for sensory, short-term, and long-term memory. For each training step, it computes the sensory memory of the output classifier and updates short-term memory based on a probabilistic decision determined by a random variable $\epsilon$ and a hyperparameter $p$. The short-term memory is subsequently consolidated into long-term memory using class-specific retention coefficients. This process ensures that the classifier's memory is continuously updated, preserving important knowledge while mitigating forgetting. The detailed steps of this memory consolidation procedure are summarized in Algorithm 2.

\begin{algorithm}[h!]
\caption{Dual-Layer Copy Weights with Re-init (D-CWR)}\label{alg:alg1}
\begin{algorithmic}
\STATE \textbf{Initialize} Initialize vector $ \boldsymbol{M^s} $, $ \boldsymbol{M^c} $, $ \boldsymbol{M^l} $ and $ \boldsymbol{P} $ to zeroes
\STATE \textbf{for} each training  time step $t$ batch samples $\boldsymbol{S}_t (\boldsymbol{X}^{in}_t\cup \boldsymbol{X}^{re}_t)$ \textbf{do}
\STATE \hspace{0.5cm} Denote $ \omega_{j} $ as the $j$-th parameter of the output classifier
\STATE \hspace{0.5cm} $\overline{\omega}$ $\leftarrow$ $\sum_{k \in S_t} \omega_{k} / |S_t|$
\STATE \hspace{0.5cm} \textbf{for} each category $j$ appearing in the batch $ \boldsymbol{S}_i $ \textbf{do}

\STATE \hspace{0.5cm} \hspace{0.5cm} $M^s_{j} \leftarrow \omega_{j} - \overline{\omega}$
\STATE \hspace{0.5cm} \hspace{0.5cm} Obtain the occurrences of class $ j $ in $ \boldsymbol{S}_t $ as $ U_j $.
\STATE \hspace{0.5cm} \hspace{0.5cm}  $\eta^c_{j} \leftarrow \frac{P_{j}}{U_{j}} $
\STATE \hspace{0.5cm} \hspace{0.5cm}  $\eta^l_{j} \leftarrow \sqrt{\frac{P_{j}}{U_{j}}}$
\STATE \hspace{0.5cm} \hspace{0.5cm} Perform one of the following operations based on a random variable $\epsilon$ and hyperparameter $p$:
\STATE \hspace{0.5cm} \hspace{0.5cm} \hspace{0.5cm} $\epsilon \ge p$: Retain $M^c_{j}$
\STATE \hspace{0.5cm} \hspace{0.5cm} \hspace{0.5cm} $\epsilon <p$: $M^c_{j} \leftarrow \frac{M^c_{j} \cdot \eta^c_{j}+ M^s_{j}}{\eta^c_{j}+1} $
\\
\STATE \hspace{0.5cm} \hspace{0.5cm} $M^l_{j} \leftarrow \frac{M^l_{j} \cdot \eta^l_{j}+M^c_{j}}{\eta^l_{j}+1}$.
\STATE \hspace{0.5cm} \hspace{0.5cm} $P_j \leftarrow P_j+U_j$.
\STATE \hspace{0.5cm}  \textbf{end for}
\STATE \hspace{0.5cm} $\boldsymbol{\omega} \leftarrow \boldsymbol{M^l}$
\STATE \textbf{end for}
\end{algorithmic}
\label{alg1}
\end{algorithm}

With the above methods, we address the issue of correlation-induced imbalance and layer-wise imbalance in OCL from two perspectives: the overall model parameters and the output classifier parameters. This results in a parameter variation balancing framework that enables network models to better adapt to non-stationary data streams. Theoretically, the ParamCC and E\&C strategies effectively capture the correlations between parameter variations and the knowledge of each task. By utilizing these correlations, 
they dynamically adjust the magnitude of parameter variations during subsequent training, thereby balancing the retention of new and old knowledge. These two strategies further enhance the performance of ER-based methods in continual learning. Not only are they highly effective in online scenarios, but they also show some benefits in offline settings. On the other hand, the D-CWR method primarily mitigates the layer-wise imbalance, proving effective in counteracting forgetting when the output classifier undergoes rapid updates.

\section{Experiments}
In this section, we first compare PVBF with competitive methods based on ER in two OCL scenarios: short task sequences and long task sequences. Next, we compare PVBF with other canonical ER-based methods in an offline CL scenario. The design of these experiments aims to validate the following performance metrics: (i) PVBF performs well across various OCL scenarios; (ii) PVBF components effectively mitigate the imbalance in parameter changes, thereby alleviating forgetting in OCL scenarios; (iii) Despite being specifically designed for OCL scenarios, PVBF still demonstrates strong performance when applied to offline CL settings.

\subsection{OCL Experiments}
\begin{figure*}[h!]
    \centering
    \begin{subfigure}{0.33\textwidth}
        \centering
        \includegraphics[width=\linewidth]{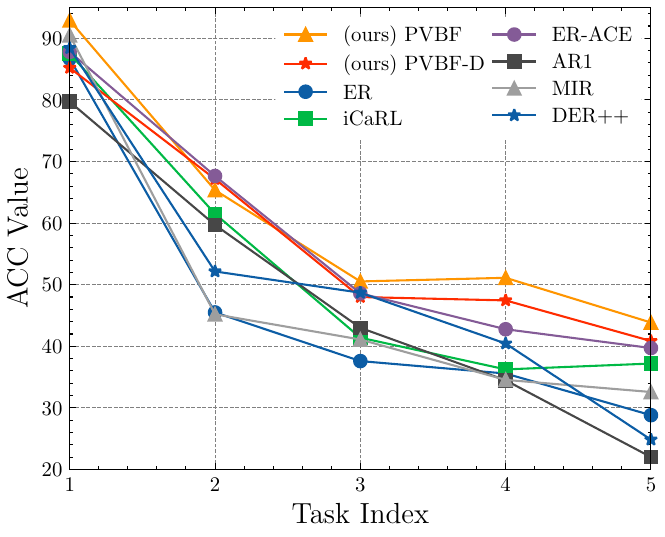}
        \caption{Cifar10 ACC results, MS=20}
        \label{fig:sub1}
    \end{subfigure}
    \begin{subfigure}{0.33\textwidth}
        \centering
        \includegraphics[width=\linewidth]{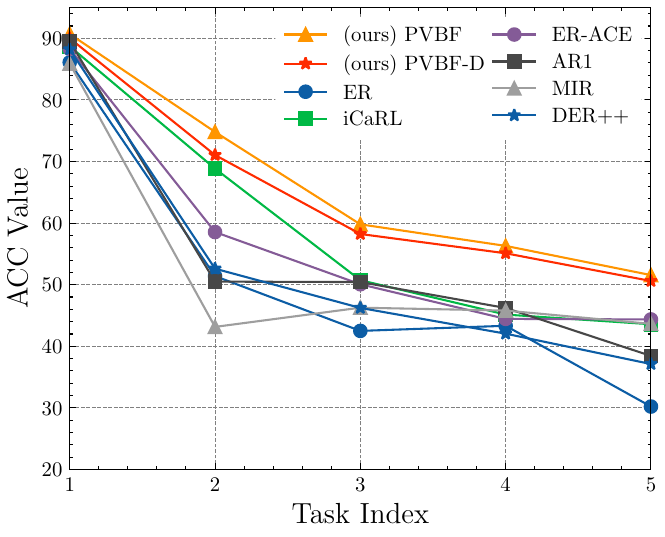}
        \caption{Cifar10 ACC results, MS=100}
        \label{fig:sub2}
    \end{subfigure}
    \begin{subfigure}{0.33\textwidth}
        \centering
        \includegraphics[width=\linewidth]{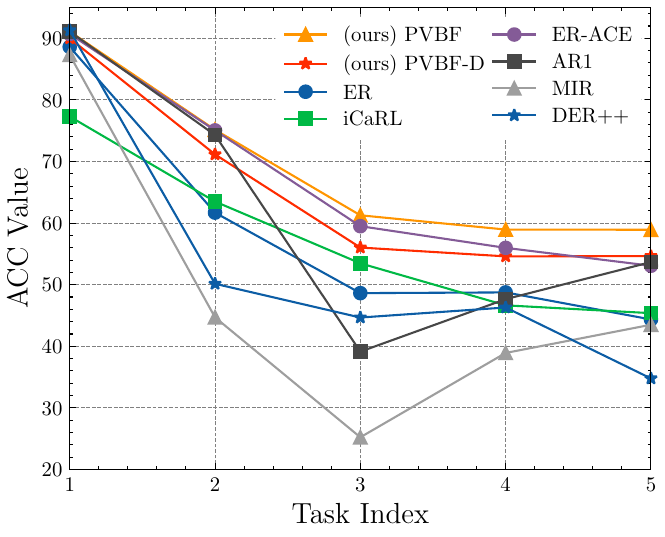}
        \caption{Cifar10 ACC results, MS=500}
        \label{fig:sub3}
    \end{subfigure}
    \caption{ACC results for short task sequence OCL experiments (PVBF-D: PVBF without D-CWR).}
    \label{fig:ACC_images1}
\end{figure*}
\begin{figure*}[h!]
    \centering
    \begin{subfigure}{0.4\textwidth}
        \centering
        \includegraphics[width=\linewidth]{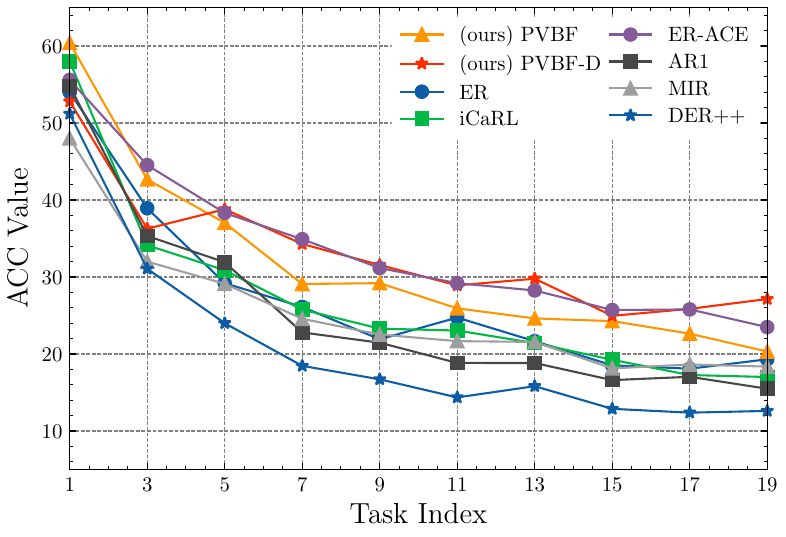}
        \caption{Cifar100 ACC results, MS=500}
        \label{fig:sub4}
    \end{subfigure}
    \begin{subfigure}{0.4\textwidth}
        \centering
        \includegraphics[width=\linewidth]{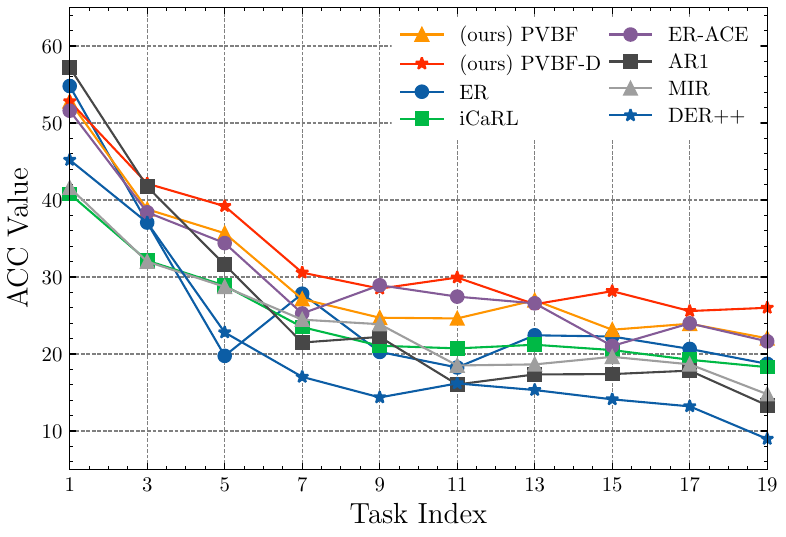}
        \caption{MiniImagenet ACC results, MS=500}
        \label{fig:sub5}
    \end{subfigure}
    \caption{ACC results for long task sequence OCL experiments (PVBF-D: PVBF without D-CWR)}
    \label{fig:ACC_images2}
\end{figure*}
We first evaluate the performance of PVBF in the OCL setting, where samples from each task are sequentially presented to the model in a one-shot manner. As such, this setup places a high demand on the model’s ability to adapt to non-stationary data streams. Only CL algorithms specifically designed for such challenges can effectively mitigate catastrophic forgetting in this setting. An effective OCL algorithm should efficiently learn the essential knowledge required to complete tasks with high accuracy while minimizing computational overhead.
\subsubsection{Datasets and Setup}

We conducted experiments to evaluate OCL for image classification tasks. As with other OCL experiments \cite{chrysakis2023online} \cite{caccia2021new}, all benchmarks were evaluated in a single-head setting. In each of the datasets mentioned below, the model performs a classification task across N classes, where N depends on the specific dataset.

We selected three representative image datasets to conduct these OCL experiments: Split Cifar10, Split Cifar100 \cite{krizhevsky2009learning}, and Split MiniImagenet \cite{vinyals2016matching}. These datasets are commonly used for evaluating OCL, with a focus on image classification tasks. The details on task division of the three adopted datasets are illustrated below:

\textbf{Split Cifar10}: This dataset consists of 60,000 images across 10 classes. It is one of the most widely adopted datasets in OCL research, often used as the standard benchmark for tasks with short task sequences. For our experiments, we partitioned the dataset into 5 tasks, each containing 2 classes.

\textbf{Split Cifar100}: The Cifar100 dataset includes 60,000 images spread across 100 classes. In our experiments, it was partitioned into 20 tasks, with each task containing 5 classes. This dataset presents a more complex and diverse challenge, making it suitable for evaluating OCL with longer task sequences.

\textbf{Split MiniImagenet}: This dataset consists of 100 classes, with each class containing 600 images. We partitioned it into 20 tasks, each with 5 classes. Like Cifar100, MiniImagenet is highly challenging for OCL due to its more diverse categories and longer task sequences.

\subsubsection{Baselines}

In our evaluation, we focus on replay-based methods due to their demonstrated effectiveness in the OCL setting, evidenced by previous studies \cite{chaudhry2019continual,aljundi2019online,caccia2021new,ji2020automatic}. Within this context, high-efficiency learning is necessary for all methods under consideration. To ensure a fair comparison, we have implemented uniform buffer management across all methods, employing Reservoir Sampling \cite{vitter1985random} to retain or discard samples. We have selected several classic and state-of-the-art OCL baselines for our analysis, including ER \cite{chaudhry2019continual}, Incremental Classifier and Representation Learning (iCaRL) \cite{rebuffi2017icarl}, Maximally Interfered Retrieval (MIR) \cite{aljundi2019online}, Dark Experience Replay (DER++) \cite{buzzega2020dark}, Experience Replay with Asymmetric Cross-Entropy (ER-ACE) \cite{caccia2021new}, and Architect-and-Replay (AR1*) \cite{lomonaco2020rehearsal}. Additionally, we include an independent and IID scenario \cite{caccia2021new} as a benchmark, where the learner is trained on the dataset with a single pass, treating all classes as if they were part of a single task. This baseline variant is designed to have a similar computational budget to that of the replay methods.
\subsubsection{Architectures and Hyperparameters} 
Like previous researches \cite{caccia2021new} \cite{buzzega2020dark} \cite{chrysakis2023online}, we use a reduced ResNet-18 network for all the datasets above, and all methods are applied to the same backbone. This choice enables us to fairly compare various methods, as these parameters directly affect the model's parameter size and the ability to classification tasks. Hyperparameters used in the experiments are presented in the following.

\textbf{Learning Rate.} The distance parameter shifts in the training process is directly affected by the learning rate. So in our experiments, for each dataset we set a fixed learning rate. Specifically, for Cifar10, we set the learning rate fixed at 0.1, for Cifar100 and MiniImagenet, we set the learning rate fixed at 0.01.

\textbf{Memory Buffer Size.} In the context of OCL scenarios, the performance of ER-based methods is closely tied to the hyperparameter of memory buffer size (MS). To ensure a fair comparison among various methods, it is necessary to evaluate their performance across different memory buffer sizes. Therefore, on the Cifar10 dataset, we varied this hyperparameter across three values: 20, 100, and 500. We conducted comparisons of selected methods under these three different settings. Conversely, on the Cifar100/MiniImageNet datasets, we fixed this hyperparameter at 500 to emphasize fairness in comparison.

\textbf{Other Hyperparameters.} Following previous studies \cite{aljundi2019online,chaudhry2018efficient,caccia2021new}, for DER++, we set the hyperparameter $\alpha=0.1$ and $\beta=0.5$. The way to realizing masking loss for ER-ACE is the same as the author proposed in \cite{buzzega2020dark}. For our PVBF, we set normalization hyperparameter $\alpha = 0.5$, $\beta = 2.0$, and the probability for D-CWR $p$ fixed at $0.9$.

\subsubsection{Metrics}

We utilized average per-task accuracy (ACC$\uparrow$, higher is better) alongside the forgetting ratio (FR$\downarrow$, lower is better) metric \cite{chaudhry2018riemannian}, to assess the continual learning capabilities of various algorithm models in OCL scenarios. The ACC metric comprehensively assesses the learning capability of algorithmic models in OCL scenarios, while the FR metric evaluates their ability to consolidate memory during the training process. Let $a_{i,j}$ be the model's accuracy on task $i$ on the test dataset after being trained on task $j$, ACC and FR are defined as
\begin{equation}
    ACC = \frac{1}{K} \sum_{i=1}^K a_{i,k}
\end{equation}
\begin{equation}
    FR = \frac{1}{K-1}  \sum_{i=1}^{K-1}{\max_{k\in \{1...K\}} a_{i,k}-a_{i,K}}
\end{equation}

\subsubsection{Short Task Sequence OCL Experiments}
\begin{table}[!htbp]

\begin{tabular*}{\textwidth}{@{\extracolsep{\fill}}lccccccc@{}}
\toprule
\hline
Method & \multicolumn{2}{c}{MS=20} & \multicolumn{2}{c}{MS=100} & \multicolumn{2}{c}{MS=500}  \\
 & ACC & FR & ACC & FR & ACC & FR  \\
\hline
\midrule
IID &  $68.0\pm1.1$ & - &  $68.0\pm1.1$ & - &  $68.0\pm1.1$ & -  \\
\hline
ER & $26.5\pm1.3$ & $50.4\pm4.2$ & $37.9\pm1.5$ & $28.1\pm1.8$ & $46.2\pm1.7$ & $18.9\pm3.6$  \\
iCaRL & $40.0\pm1.1$ & $38.5\pm1.6$ & $42.4\pm1.1$ & $32.3\pm1.5$ & $44.1\pm1.0$ & $30.6\pm1.4$  \\
MIR & $28.8\pm1.4$ & $46.7\pm2.5$ & $46.7\pm0.9$ & $\underline{18.8}\pm2.5$ & $47.0\pm2.0$ & $\boldsymbol{9.1}\pm2.8$ \\
DER++ & $30.8\pm1.8$ & $33.7\pm4.3$ & $39.8\pm1.3$ & $23.3\pm2.1$ & $47.8\pm1.6$ & $13.9\pm2.8$ \\
ER-ACE & $37.7\pm0.8$ & $28.8\pm2.7$ & $\underline{47.4}\pm1.2$ & $20.0\pm2.6$ & $53.3\pm1.4$ & $15.1\pm1.7$  \\
AR1* & $24.5\pm1.4$ & $75.9\pm1.8$ & $38.9\pm1.8$ & $53.0\pm2.5$ & $51.6\pm2.5$ & $36.4\pm3.7$ \\
\hline
(ours) PVBF w/o D-CWR & $\underline{37.9}\pm1.0$ & $\underline{26.6}\pm2.0$ & $47.1\pm1.3$ & $19.0\pm1.7$ & $\underline{54.8}\pm1.7$ & $12.7\pm2.0$ \\
(ours) PBVF & \textbf{$\boldsymbol{41.2}\pm0.8$} & $\boldsymbol{24.0}\pm2.0$ & $\boldsymbol{50.8}\pm1.4$ & $\boldsymbol{18.1}\pm1.8$ & $\boldsymbol{59.2}\pm1.8$ & $\underline{10.6}\pm1.0$  \\
\bottomrule
\hline
\end{tabular*}
\caption{Short task sequence experiment results on split Cifar10. We conducted experiments in three scenarios: $MS=20$, $MS=100$, $MS=500$. All entries are 95\%-confidence intervals over 15 runs.}
\label{tab:short}
\end{table}
First as a standard short task sequence setting \cite{caccia2021new}, we apply our methods on Cifar10 datasets, along with the baselines. The results are shown in Table \ref{tab:short} and Fig.~\ref{fig:ACC_images1}.

In experiments on Cifar10 under standard OCL scenarios, varying memory buffer sizes reveal distinct impacts on performance metrics. The IID method reveals the upper bound of learning accuracy for network models on this dataset (where no forgetting issue exists). As in Table \ref{tab:short}, In scenarios with $MS$=20, $MS$=100, and $MS$=500, our PBVF demonstrates a significant accuracy advantage over other baselines, achieving an average $39.2\%$ improvement in ACC compared to the ER method. Furthermore, our approach is able to further reduce the FR metric based on ER-ACE and AR1*, and has the lowest FR metrics in $MS=20$ and $MS=100$ settings, demonstrating more effective consolidating memory of past knowledge.

The results of the short task sequence experiments not only demonstrate that our PVBF exhibits strong adaptability in this setting, but also validate our hypothesis regarding the parameter variation imbalance. By comparing the PVBF without D-CWR to the ER-ACE method, we observe that the PVBF without D-CWR consistently achieves better performance across $MS=20$, and $MS=500$ settings, and lower FR in all settings, indicating that the E\&C strategy alleviates the correlation-induced imbalance. Furthermore, when the D-CWR strategy is applied, the PVBF framework shows significantly lower FR and higher ACC, which can be attributed to the severe layer-wise imbalance in the short-task sequence OCL setting, and our D-CWR strategy effectively mitigates this imbalance.

\subsubsection{Long Task Sequence OCL Experiments }
\begin{table}[!htbp]

\centering
\begin{tabular*}{\textwidth}{@{\extracolsep{\fill}}lccccccc@{}}
\toprule
\hline
Method & \multicolumn{2}{c}{Cifar100} & \multicolumn{2}{c}{MiniImagenet}   \\
 & ACC & FR & ACC & FR   \\
\hline
\midrule
IID & $28.3\pm0.9$& -  & $24.3\pm1.6$& -  \\
\hline
ER & $17.5\pm0.5$& $50.4\pm0.7$ & $18.1\pm0.6$& $40.8\pm1.1$  \\
iCaRL & $17.1\pm0.3$& $22.1\pm0.3$  & $16.8\pm0.3$& $15.8\pm0.6$  \\
MIR & $18.1\pm0.5$& $47.9\pm0.7$  & $18.9\pm0.6$& $38.3\pm0.9$ \\
DER++ & $10.9\pm0.6$& $61.7\pm0.7$  & $10.3\pm0.7$& $51.6\pm1.0$ \\
ER-ACE & $\underline{24.0}\pm0.7$& $\boldsymbol{10.1}\pm0.7$  & $\underline{23.2}\pm0.7$& $\boldsymbol{8.9}\pm0.8$  \\
AR1* & $14.6\pm0.4$& $58.1\pm0.7$  & $15.7\pm0.5$& $47.9\pm0.6$ \\
\hline

(ours) PVBF w/o D-CWR & $\boldsymbol{25.8}\pm0.7$& $\underline{10.3}\pm0.8$  & $\boldsymbol{23.7}\pm0.8$& $\underline{9.2}\pm0.9$  \\
(ours) PVBF  & $21.7\pm0.7$& $11.4\pm0.6$  & $20.9\pm0.4$& $9.3\pm0.5$ 

\\
\bottomrule
\hline
\end{tabular*}
\caption{Long task sequence experiment results on split Cifar100 and split MiniImagenet. All entries are 95\%-confidence intervals over 15 runs.}
\label{tab:long}
\end{table}
Similar experiments are conducted with the split Cifar100 and split MiniImagenet dataset with 20 tasks. The results are shown in Table \ref{tab:long} and Fig.~\ref{fig:ACC_images2}.

The experimental results demonstrate that our PVBF without D-CWR outperforms all baseline methods on both the Cifar100 and Miniimagenet datasets. As in Table \ref{tab:long}, particularly on the split Cifar100 dataset, the PVBF without D-CWR shows a $47\%$ improvement in ACC over the ER method, approaching the model accuracy upper bound indicated by the IID method. On the split Miniimagenet dataset, the PVBF without D-CWR also performs well, showing improvements over the previously best-performing ER-ACE method. Additionally, the FR metric for the PVBF without D-CWR is relatively low in both datasets, very close to the best ER-ACE results, but for the ACC metric, PVBF without D-CWR has a clear advantage. This confirms that the evaluation of parameter importance and overall gradient correction in the PVBF without D-CWR are effective for long task sequences in the OCL scenario.
\begin{figure*}[h!]
    \centering
    \begin{subfigure}{0.24\textwidth}
        \centering
        \includegraphics[width=\linewidth]{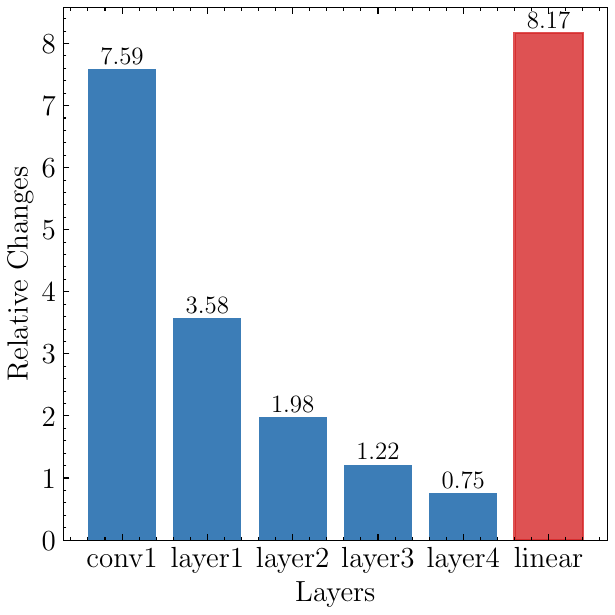}
        \caption{Task 1}
        \label{fig:sub1}
    \end{subfigure}
    \begin{subfigure}{0.24\textwidth}
        \centering
        \includegraphics[width=\linewidth]{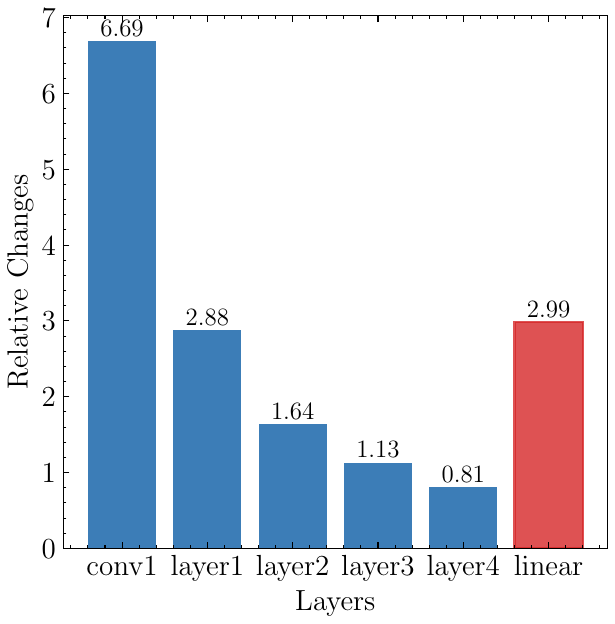}
        \caption{Task 6}
        \label{fig:sub2}
    \end{subfigure}
    \begin{subfigure}{0.24\textwidth}
        \centering
        \includegraphics[width=\linewidth]{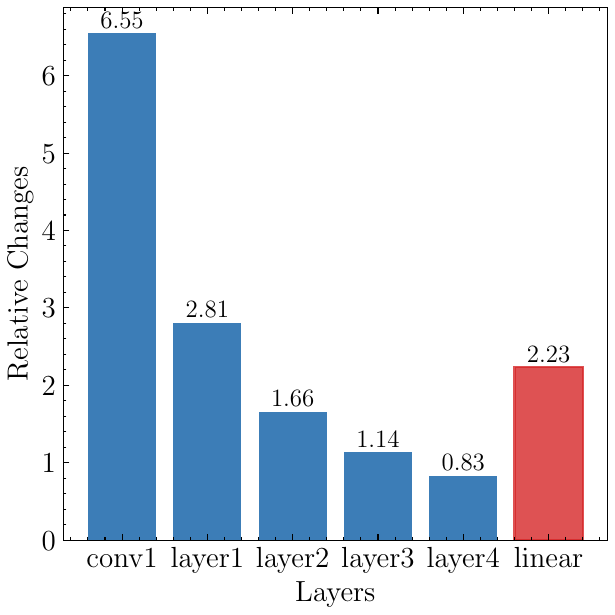}
        \caption{Task 11}
        \label{fig:sub3}
    \end{subfigure}
    \begin{subfigure}{0.24\textwidth}
        \centering
        \includegraphics[width=\linewidth]{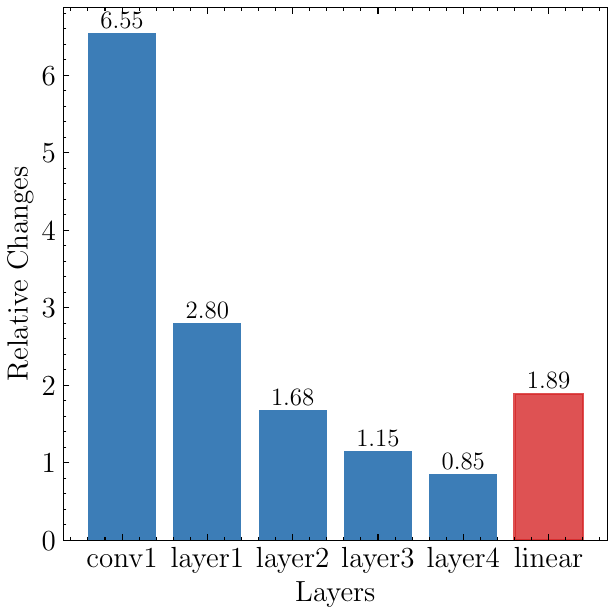}
        \caption{Task 16}
        \label{fig:sub4}
    \end{subfigure}
    \caption{Average relative changes in parameters across different layers in a long task sequence OCL experiment on MiniImagenet applying PVBF.}
    \label{fig:f5_images}
\end{figure*}

The OCL settings with long task sequences reveal significant differences in parameter variation imbalance compared to those with short task sequences. D-CWR may reduce the performance of PVBF, and the AR1* method, which only uses output classifier optimization, also performs poorly. We hypothesize that this is due to the increased complexity and more frequent changes in tasks within the long task sequence scenario, which requires the output classifier to update more rapidly. In this scenario, as illustrated in Fig.~\ref{fig:f5_images}, the relative changes in the parameters of the output classifier (linear layer) gradually decrease as the number of tasks increases, ultimately leading to a diminished effect of layer-wise imbalance. In other words, in these settings, correlation-induced imbalance is the primary contributor to parameter variation imbalance. Our PVBF without D-CWR primarily focuses on optimizing parameter dependency imbalance, which allows it to achieve performance close to that of the IID method.
\subsubsection{Ablation Study}
Based on the results of the ablation study presented in Table \ref{tab:ablation}, several key observations can be made regarding the performance of PVBF on the split Cifar10 dataset. The results of the ablation experiments demonstrate significant impacts of removing different modules on the PVBF. Removing the E\&C  and D-CWR modules decreased the ACC from \(59.2\%\) to \(55.9\%\) and \(54.8\% \), indicating their substantial contributions to accuracy. Simultaneously, the FR values for these modules increased from \(10.6\%\) to \(13.3\%\) and \(12.7\% \), illustrating their effectiveness in reducing forgetting. This implies that it is essential to correct prediction bias both holistically and specifically in the output classifier. Removing the ACE module caused the ACC to drop from \(59.2\% \)  to \(51.3\% \), and FR to increase from $10.6$ to $41.8$, highlighting the importance of combining ACE into PVBF to mitigate catastrophic forgetting. 
\begin{table}[htbp]
\centering
\begin{tabular}{lcccc}
\hline
\textbf{PVBF w/o} & \textbf{E\&C} & \textbf{D-CWR} & \textbf{ACE} & \textbf{-} \\
\hline
\textbf{ACC} & $55.9\pm1.1$ & $54.8\pm1.7$ & $51.3\pm2.7$ & $59.2\pm1.8$ \\
\textbf{FR} & $13.3\pm1.6$ & $12.7\pm2.0$ & $41.8\pm3.1$ & $10.6\pm1.0$ \\
\hline
\end{tabular}
\caption{Ablation experiment results on split Cifar10. All entries are 95\%-confidence intervals over 15 runs.}
\label{tab:ablation}
\end{table}

\begin{table}[!h]

\begin{tabular*}{\textwidth}{@{\extracolsep{\fill}}lccccccc@{}}
\toprule

Method & \multicolumn{2}{c}{MS=200} & \multicolumn{2}{c}{MS=500} & \multicolumn{2}{c}{MS=5120}  \\
 & ACC & FR & ACC & FR & ACC & FR  \\
\midrule
\multicolumn{7}{c}{Class-IL}\\
\hline
SGD &  $19.62\pm0.05$ & $96.39\pm0.12$ &  $19.62\pm0.05$ & $96.39\pm0.12$  &  $19.62\pm0.05$ & $96.39\pm0.12$   \\
ER & $44.79\pm1.86$ & $61.24\pm2.62$ & $57.74\pm0.27$ & $45.35\pm0.07$ & $82.57\pm0.52$ & $13.99\pm1.12$  \\
GEM & $25.54\pm0.76$ & $82.61\pm1.60$ & $26.20\pm2.26$ & $74.31\pm4.62$ & $25.26\pm3.46$ & $75.27\pm4.41$ \\
iCaRL & $49.02\pm3.20$ & $28.72\pm0.49$ & $47.55\pm3.95$ & $25.71\pm1.10$ & $55.07\pm1.55$ & $24.94\pm0.14$ \\
FDR & $30.91\pm2.74$ & $86.40\pm0.67$ & $28.71\pm3.23$ & $85.62\pm0.36$ & $19.70\pm0.07$ & $96.64\pm0.19$  \\
GSS & $39.07\pm5.59$ & $75.25\pm4.07$ & $49.73\pm4.78$ & $62.88\pm2.67$ & $67.27\pm4.27$ & $58.11\pm9.12$ \\
HAL &$32.36\pm2.70$ & $69.11\pm4.21$ & $41.79\pm4.46$ & $62.21\pm4.34$ & $59.12\pm4.41$ & $27.19\pm7.53$ \\
DER++ &$64.88\pm1.17$ & $32.59\pm2.32$ & $\textbf{72.70}\pm1.36$ & $22.38\pm4.41$ & $\textbf{85.24}\pm0.49$ & $7.27\pm0.84$ \\
ER-ACE &$63.48\pm1.24$ & $\textbf{16.43}\pm1.65$ & $69.95\pm1.18$ & $\textbf{12.58}\pm0.79$ & $83.06\pm0.62$ & $\underline{6.14}\pm0.88$ \\
(ours) PBVF & \textbf{$\boldsymbol{65.27}\pm1.18$} & $\underline{22.11}\pm1.95$ & $\underline{71.30}\pm1.01$ & $\underline{14.49}\pm1.25$ & $\underline{83.45}\pm0.67$ & $\textbf{6.12}\pm0.61$  \\
\hline
\multicolumn{7}{c}{Task-IL}\\
\end{tabular*}
\begin{tabular*}{\textwidth}{@{\extracolsep{\fill}}lccccccc@{}}
\hline

SGD &  $61.02\pm3.33$ & $46.24\pm2.12$ &  $61.02\pm3.33$ & $46.24\pm2.12$  &  $61.02\pm3.33$ & $46.24\pm2.12$   \\
ER & $91.19\pm0.94$ & $7.08\pm0.64$ & $93.61\pm0.27$ & $3.54\pm0.35$ & $96.98\pm0.17$ & $\textbf{0.27}\pm0.06$  \\
GEM & $90.44\pm0.94$ & $9.27\pm2.07$ & $92.16\pm0.69$ & $9.12\pm0.21$ & $25.26\pm3.46$ & $6.91\pm2.33$ \\
iCaRL & $88.99\pm2.13$ & $\textbf{2.63}\pm3.48$ & $88.22\pm2.62$ & $\textbf{2.66}\pm2.47$ & $92.23\pm0.84$ & $1.59\pm0.57$ \\
FDR & $91.01\pm0.68$ & $7.36\pm0.03$ & $93.29\pm0.59$ & $4.80\pm0.00$ & $94.32\pm0.97$ & $1.93\pm0.48$  \\
GSS & $88.80\pm2.89$ & $8.56\pm1.78$ & $91.02\pm1.57$ & $7.73\pm3.99$ & $94.19\pm1.15$ & $7.71\pm2.31$ \\
HAL &$82.51\pm3.20$ & $12.26\pm0.02$ & $84.54\pm2.36$ & $5.41\pm1.10$ & $88.51\pm3.32$ & $5.21\pm0.50$ \\
DER++ &$91.92\pm0.60$ & $5.16\pm0.21$ & $93.88\pm0.50$ & $4.66\pm1.15$ & $96.12\pm0.21$ & $1.18\pm0.19$ \\
ER-ACE &$\underline{92.74}\pm1.18$ & $5.59\pm0.90$ & $\underline{94.35}\pm0.92$ & $3.58\pm0.72$ & $\textbf{97.25}\pm0.48$ & $0.78\pm0.16$ \\
(ours) PBVF & \textbf{$\boldsymbol{93.96}\pm1.06$} & $\underline{4.80}\pm1.03$ & $\boldsymbol{95.62}\pm0.71$ & $\underline{2.68}\pm0.49$ & $\underline{97.12}\pm0.50$ & $\underline{0.76}\pm0.12$  \\
\bottomrule
\end{tabular*}
\caption{Offline experiment results on split Cifar10. $MS$ represents the memory buffer size. We conducted experiments in three scenarios: $MS=200$, $MS=500$, $MS=5120$. Results by SGD, ER, GEM, iCaRL, FDR, GSS, HAL, and DER++ are cited from \cite{buzzega2020dark}, all other entries are 95\%-confidence intervals over 15 runs.}
\label{tab:offline}
\end{table}
\subsection{Offline CL Experiments}
To compare with more ER-based CL methods designed for offline scenarios and to further validate the stable performance of PVBF in more CL settings, we applied PVBF and other baselines to the same offline CL setting. The offline setting typically includes two specific learning modes: class-incremental  (class-IL) and task-incremental (task-IL). The primary difference between these two learning modes lies in the evaluation process. In the class-IL learning mode, task identifiers are not provided to the model, and the classifier must accurately predict the class of validation samples among all seen classes. In contrast, for the task-IL learning mode, task identifiers are provided to the model, and the classifier only needs to classify samples within the corresponding task. In the offline CL setting, we use the same evaluation metrics as in the OCL scenario.

\subsubsection{Datasets and Setup}
In reference to the standard research on offline continual learning (CL) in \cite{buzzega2020dark}, we conduct experiments on the widely used Split-Cifar10 dataset, addressing both class-IL and task-iL scenarios. These two scenarios respectively assess the model's memory capabilities for the entire dataset and for individual tasks, making them highly representative in the context of offline CL. Similar to the OCL scenario, we divide the Split-Cifar10 dataset into five tasks, with each task requiring the model to classify samples from two distinct classes.
\subsubsection{Baselines}
Regarding the selection of baselines, we introduce seven existing ER-based methods from offline CL experiments (\textbf{ER} \cite{riemer2018learning}, \textbf{GEM} \cite{lopez2017gradient}, \textbf{iCaRL} \cite{rebuffi2017icarl}, \textbf{FDR} \cite{benjaminmeasuring}, \textbf{GSS} \cite{aljundi2019gradient}, \textbf{HAL} \cite{chaudhry2021using}, and \textbf{DER++}), as well as \textbf{ER-ACE}, which has shown strong performance in OCL scenarios. Similar to the OCL scenario, for the nine methods, including PVBF, we employ reservoir sampling for data sampling. To ensure a fair comparison across methods, we set the same memory buffer size for all methods within the same experimental setting. Simultaneously, to highlight the memory advantages offered by continual learning algorithms, we also include experimental results for the SGD method. This approach does not utilize any experience replay, and instead performs standard stochastic gradient descent training on each task sequentially, with the same number of training epochs as the other algorithms.

\subsubsection{Training}
During training, as we adopted part of the offline CL experimental data from prior work, we ensured consistency across all conditions that influence model performance. Specifically, we adhered to the following characteristics of the Mammoth experiment framework \cite{boschini2022class}, which is designed for offline CL experiments: (i) For all training processes, we used the same SGD optimizer with a fixed batch size of 32 and fixed epochs 50; (ii) All methods were applied to a common, unpretrained ResNet-18 based backbone; (iii) Experiments were conducted across three different memory buffer sizes: 200, 500, and 5120. For the seven classic baselines (ER, GEM, iCaRL, FDR, GSS, HAL, and DER++), we referenced their optimal performance results under previously validated hyperparameter settings for this configuration. For ER-ACE and PVBF, we conducted experiments with a fixed learning rate of 0.03. The hyperparameter settings for PVBF were consistent with those used in the OCL setting, and all experimental results are fully reproducible.

\subsubsection{Results}
As shown in Table \ref{tab:offline}, experiments conducted in the offline CL settings demonstrate that PVBF exhibits stable performance across both class-IL and task-IL scenarios. In the class-IL setting, PVBF achieves the highest ACC and the second-best FR when the memory buffer size $MS = 200$. For $MS = 500$ and $MS = 5120$, PVBF consistently shows the second-highest ACC and either the lowest or second-lowest FR. In the task-IL scenario, PVBF achieves optimal performance for $MS = 200$ and $MS = 500$, surpassing all baselines. For $MS = 5120$, PVBF's performance is on par with ER-ACE. Overall, across both scenarios, PVBF demonstrates a competitive advantage over existing state-of-the-art methods in the offline CL setting. Notably, particularly in settings with a small memory buffer size and in task-IL scenarios, PVBF retains a clear advantage over DER++, which also demonstrates excellent performance in the offline CL setting.
\subsubsection{Analysis}
In the offline CL setting, compared to the OCL setting, the model is able to undergo more extensive training, which alleviates the issue of parameter variation imbalance. Particularly, in the case of $MS = 5120$, the model can store a sufficient number of samples during training, allowing it to effectively revisit knowledge from previous tasks. In this scenario, the issue of parameter variation imbalance is mitigated under the conditions of ample sample and training data. As a result, the experimental outcomes of ER-based methods eventually converge with those of the ER method. When the number of samples that the model can store is smaller, the issue of parameter variation imbalance becomes more pronounced. In such cases, PVBF can alleviate this issue to some extent, leading to better performance compared to other methods. In the task-IL scenario, the correlation-induced imbalance correction strategy employed by PVBF enables the model to retain knowledge of specific tasks more vividly, which contributes to PVBF's superior performance.

\section{Conclusion}

Mitigating catastrophic forgetting in OCL requires addressing the prediction bias caused by parameter variation imbalance. To solve this issue, this paper proposes a Parameter Variation Balancing Framework (PVBF), which mitigates the correlation-induced imbalance and layer-wise imbalance respectively. The paper validates PVBF through classification experiments on both short and long task sequence OCL settings, as well as offline settings. The results show that PVBF achieves an average accuracy improvement of 31\%-47\% over the ER method. On the MiniImageNet dataset, it attains 97.5\% of the IID method's accuracy using only 500 replay samples. In offline CL settings, PVBF consistently outperforms classical CL methods, demonstrating its significant advantages.
\appendix
\section{Relative Changes Analysis of Different Standardization Methods}
\label{app1}
\begin{figure*}[h!]
    \centering
    \begin{subfigure}{0.24\textwidth}
        \centering
        \includegraphics[width=\linewidth]{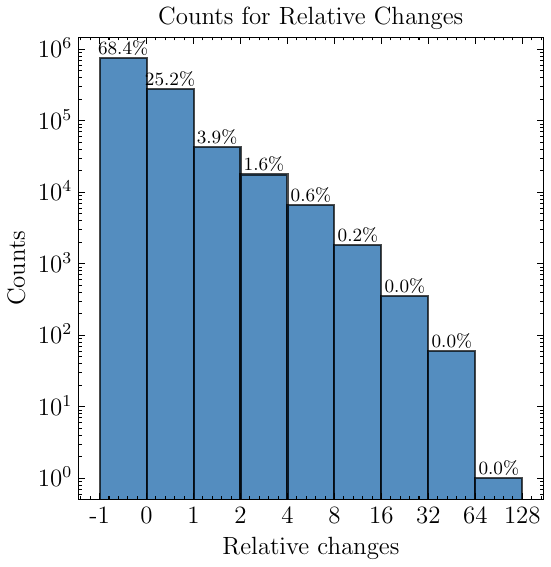}
        \caption{Task 1}
        \label{fig:sub1}
    \end{subfigure}
    \begin{subfigure}{0.24\textwidth}
        \centering
        \includegraphics[width=\linewidth]{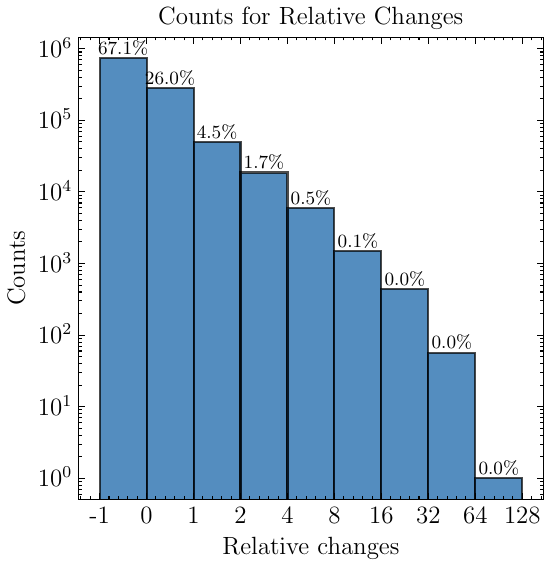}
        \caption{Task 2}
        \label{fig:sub2}
    \end{subfigure}
    \begin{subfigure}{0.24\textwidth}
        \centering
        \includegraphics[width=\linewidth]{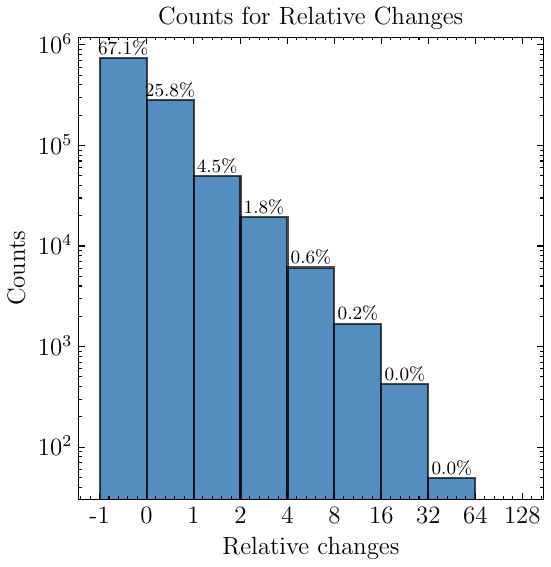}
        \caption{Task 3}
        \label{fig:sub3}
    \end{subfigure}
    \begin{subfigure}{0.24\textwidth}
        \centering
        \includegraphics[width=\linewidth]{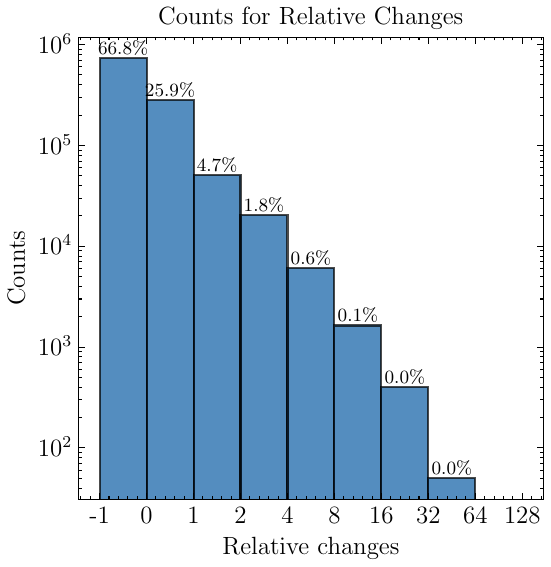}
        \caption{Task 4}
        \label{fig:sub4}
    \end{subfigure}
    \caption{Neuron counts for different relative changes $\delta_{m,k}^{'}$ (ZS method)}
    \label{fig:ZS1}
\end{figure*}
\begin{figure*}[h!]
    \centering
    \begin{subfigure}{0.24\textwidth}
        \centering
        \includegraphics[width=\linewidth]{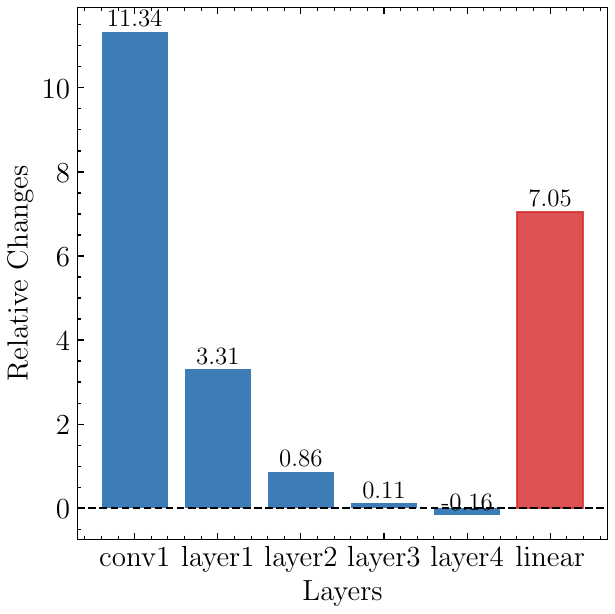}
        \caption{Task 1}
    \end{subfigure}
    \begin{subfigure}{0.24\textwidth}
        \centering
        \includegraphics[width=\linewidth]{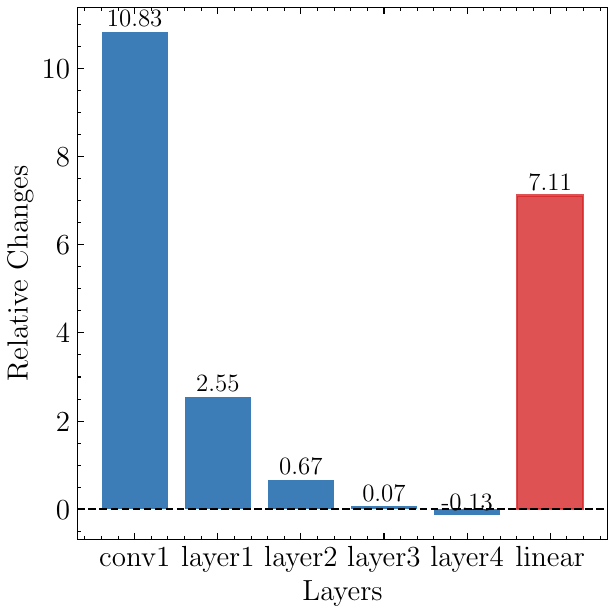}
        \caption{Task 2}

    \end{subfigure}
    \begin{subfigure}{0.24\textwidth}
        \centering
        \includegraphics[width=\linewidth]{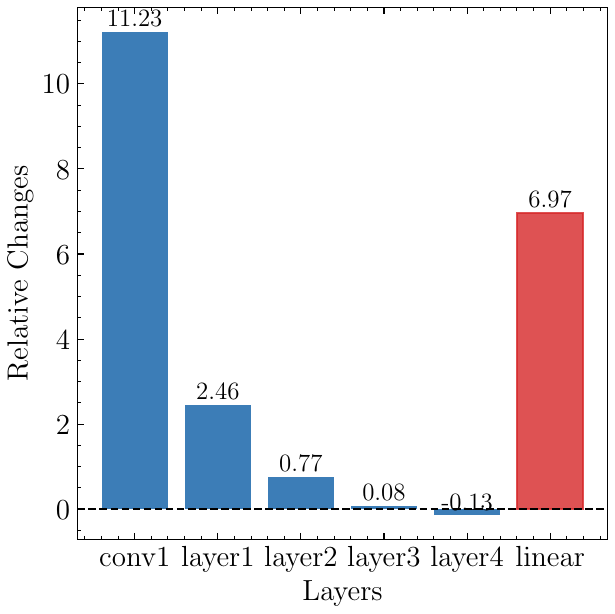}
        \caption{Task 3}

    \end{subfigure}
    \begin{subfigure}{0.24\textwidth}
        \centering
        \includegraphics[width=\linewidth]{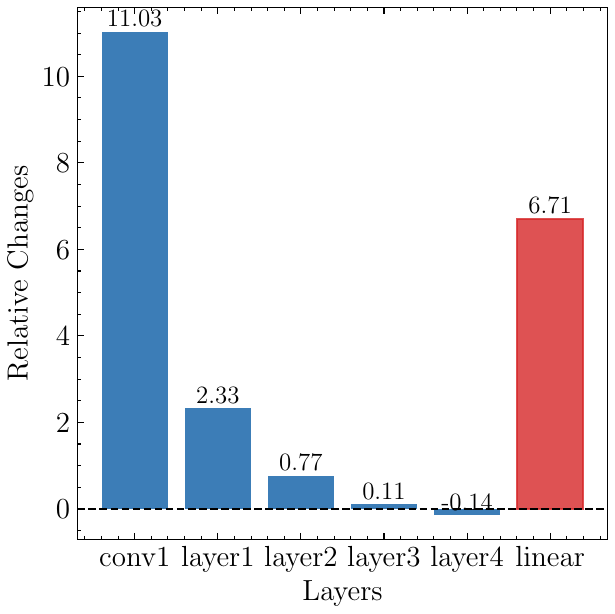}
        \caption{Task 4}

    \end{subfigure}
    \caption{Average relative changes in parameters across different layers (ZS method)}
    \label{fig:ZS2}
\end{figure*}
\begin{figure*}[h!]
    \centering
    \begin{subfigure}{0.24\textwidth}
        \centering
        \includegraphics[width=\linewidth]{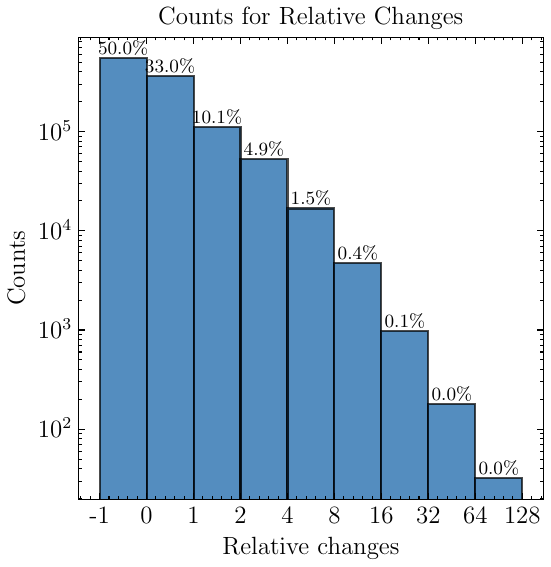}
        \caption{Task 1}
        \label{fig:sub1}
    \end{subfigure}
    \begin{subfigure}{0.24\textwidth}
        \centering
        \includegraphics[width=\linewidth]{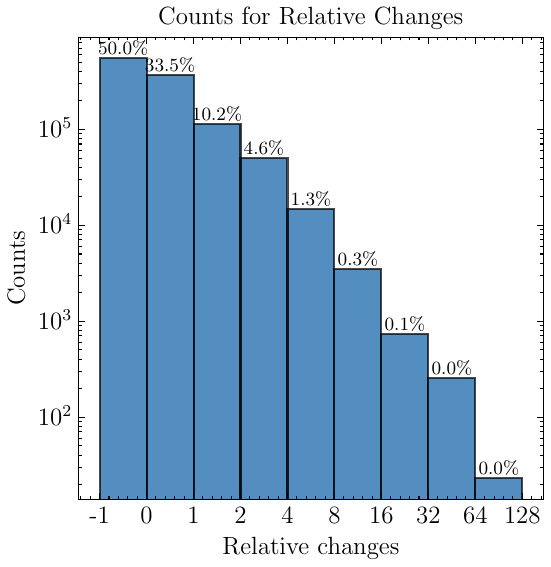}
        \caption{Task 2}
        \label{fig:sub2}
    \end{subfigure}
    \begin{subfigure}{0.24\textwidth}
        \centering
        \includegraphics[width=\linewidth]{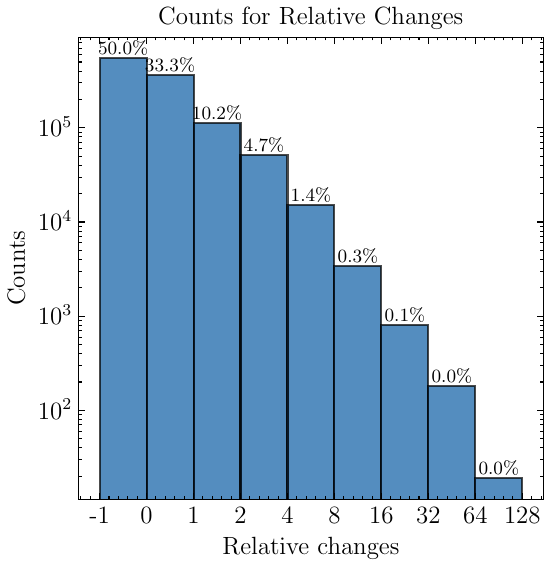}
        \caption{Task 3}
        \label{fig:sub3}
    \end{subfigure}
    \begin{subfigure}{0.24\textwidth}
        \centering
        \includegraphics[width=\linewidth]{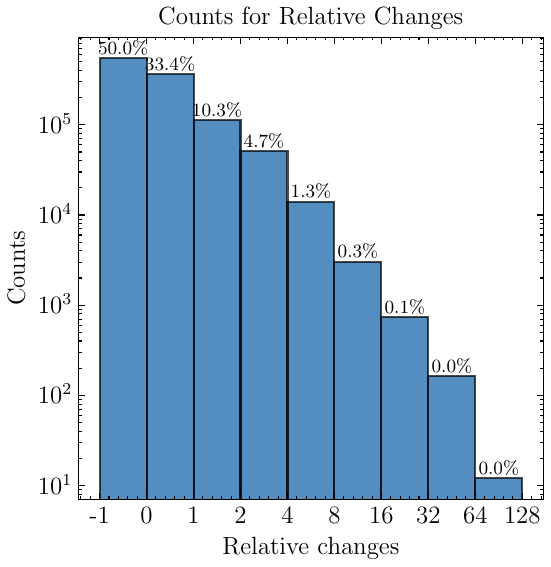}
        \caption{Task 4}
        \label{fig:sub4}
    \end{subfigure}
    \caption{Neuron counts for different relative changes $\delta_{m,k}^{'}$ (RS method)}
    \label{fig:RS1}
\end{figure*}
\begin{figure*}[h!]
    \centering
    \begin{subfigure}{0.25\textwidth}
        \centering
        \includegraphics[width=\linewidth]{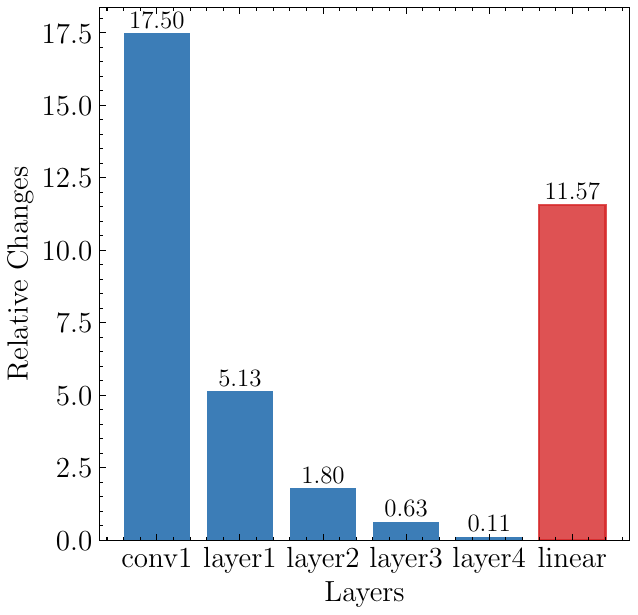}
        \caption{Task 1}
    \end{subfigure}
    \begin{subfigure}{0.23\textwidth}
        \centering
        \includegraphics[width=\linewidth]{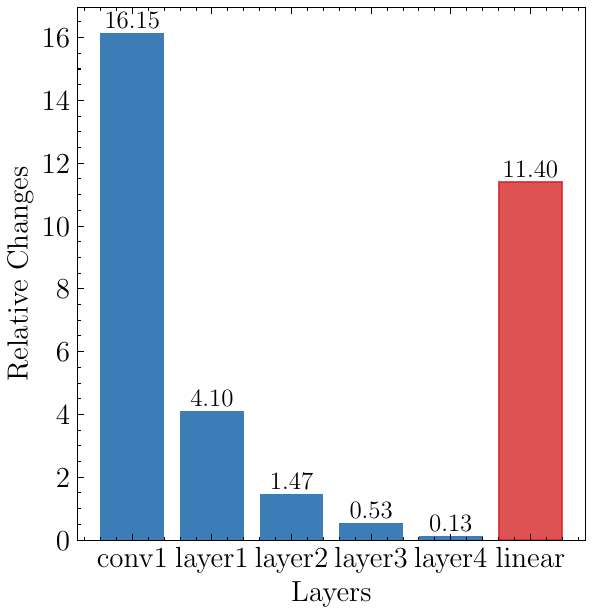}
        \caption{Task 2}

    \end{subfigure}
    \begin{subfigure}{0.23\textwidth}
        \centering
        \includegraphics[width=\linewidth]{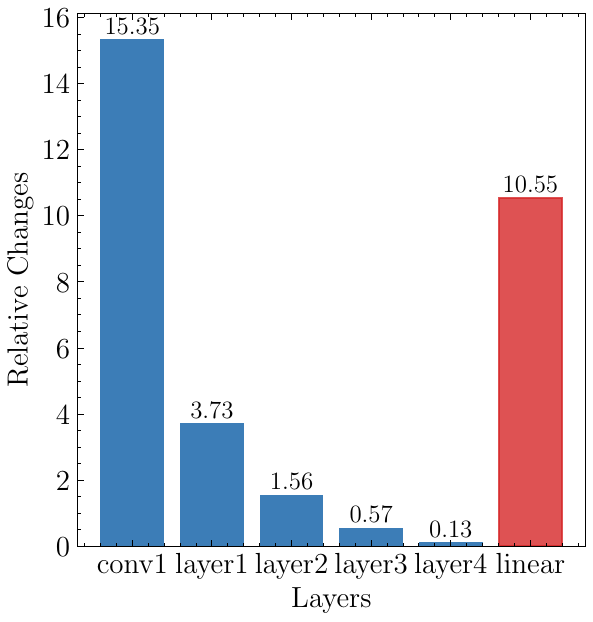}
        \caption{Task 3}

    \end{subfigure}
    \begin{subfigure}{0.24\textwidth}
        \centering
        \includegraphics[width=\linewidth]{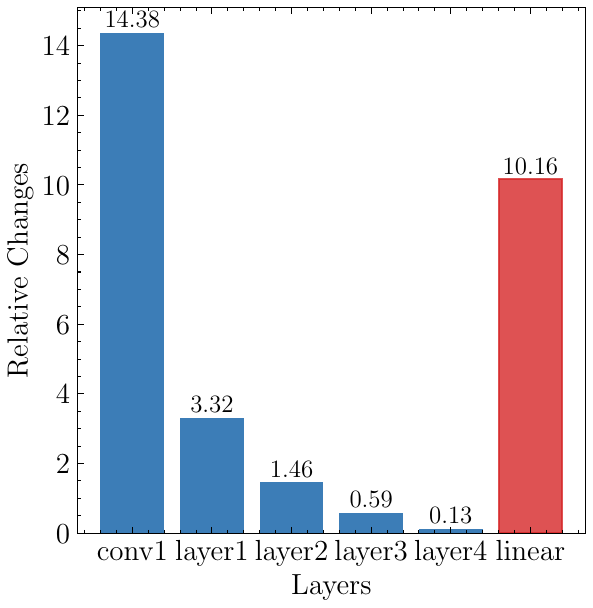}
        \caption{Task 4}

    \end{subfigure}
    \caption{Average relative changes in parameters across different layers in average (RS method)}
    \label{fig:RS2}
\end{figure*}

We also use the ZS and RS method for obtaining $\delta_{m,k}^{'}$, we recorded the relative changes at the end of $1\sim 4$ task in a sequence of short tasks (a total of 5 tasks) during continual training on Cifar10 \cite{krizhevsky2009learning} with ER method \cite{chaudhry2019continual} using a backbone of reduced-Resnet18 \cite{aljundi2019online}. 

From the results in Fig.~\ref{fig:ZS1} using ZS normalization, we observe that in all tasks, 66\%-68\% of the parameters exhibit relative changes below the mean, and over 90\% of the parameters have relative changes within one standard deviation. This further supports our hypothesis regarding correlation-induced imbalance, where only a small subset of parameters show a strong correlation within a task. Similarly, the results in Fig.~\ref{fig:RS1} using RS normalization corroborate this observation, with the number of parameters exhibiting relative changes greater than 2 being less than 7\% after normalization.

Similarly, the results shown in Fig.~\ref{fig:ZS2}, which present the average relative changes of parameters at different layers using ZS normalization, and those in Fig.~\ref{fig:RS2}, which show the corresponding results with RS normalization, further support our hypothesis regarding layer-wise imbalance. In both cases, the final linear layer (i.e., the output classifier) exhibits significantly higher average relative changes in parameters compared to the adjacent layers.
\section{OCL Experiments Results of Different Standardization Methods}
\label{app2}

In this section, we present the experimental results of the PVBF framework under different standardization methods. The results are evaluated across multiple task sequences with varying sequence lengths and datasets. We report the ACC and FR for each method.

\textbf{Short Task Sequence Results}. Table \ref{tab:short2} summarizes the results for the short task sequence experiment conducted on the split Cifar10 dataset. The results highlight the performance of PVBF with three standardization methods: RR, ZS, and RS. Notably, the PBVF-RR method exhibits the highest ACC when $MS=500$ and lowest FR when $MS=20$. The PBVF-ZS method achieves the best ACC at 51.6 ± 1.0 when $MS=100$, and lowest FR when $MS=100$ and $MS=500$. The PVBF-RS method achieves best ACC when $MS=20$. From the results of the short task sequence, both PVBF-RR and PVBF-ZS demonstrate comparable performance, outperforming PVBF-RS.

\textbf{Long Task Sequence Results}. Table \ref{tab:long2} presents the results from the long task sequence experiment conducted on the split Cifar100 and split MiniImagenet datasets. Again, we evaluate the performance of PVBF with three standardization methods: RR, ZS, and RS. In the case of Cifar100, PVBF-RS achieves the highest ACC of 22.8, outperforming both PVBF-RR and PVBF-ZS, and PVBF-ZS achieves the lowest FR. On the MiniImagenet dataset, PVBF-RS achieves the highest ACC of 21.8, while PVBF-RR shows a lower forgetting rate of 9.8. In contrast to the short task sequence results, PVBF-RS outperforms the other two methods in the long task sequence.

Considering both OCL scenarios, all three standardization methods exhibit similar performance on the OCL task. From a practical perspective, RR has the lowest computational complexity among the three methods. Therefore, we primarily use the RR method in the main discussions.
\begin{table}[!htbp]  

\begin{tabular*}{\textwidth}{@{\extracolsep{\fill}}lccccccc@{}}
\toprule

Method & \multicolumn{2}{c}{MS=20} & \multicolumn{2}{c}{MS=100} & \multicolumn{2}{c}{MS=500}  \\
 & ACC & FR & ACC & FR & ACC & FR  \\

\midrule
PBVF-RR & \textbf{$41.2\pm0.8$} & $\boldsymbol{24.0}\pm2.0$ & $50.8\pm1.4$ & $18.1\pm1.8$ & $\boldsymbol{59.2}\pm1.8$ & $10.6\pm1.0$  \\
PBVF-ZS & \textbf{$40.4\pm1.6$} & $24.4\pm2.0$ & $\boldsymbol{51.6}\pm1.0$ & $\boldsymbol{14.8}\pm1.2$ & $59.0\pm2.1$ & $\textbf{10.1}\pm2.5$  \\
PBVF-RS & \textbf{$\boldsymbol{42.0}\pm1.1$} & $24.2\pm2.2$ & $49.5\pm1.6$ &$18.2\pm1.8$ & $56.3\pm1.8$ & $13.3\pm1.2$  \\
\bottomrule
\hline
\end{tabular*}
\caption{Short task sequence experiment results on split Cifar10 using PVBF with different standardization methods. All entries are 95\%-confidence intervals over 15 runs.}
\label{tab:short2}
\end{table}

\begin{table}[!htbp]

\centering
\begin{tabular*}{\textwidth}{@{\extracolsep{\fill}}lccccccc@{}}
\toprule

Method & \multicolumn{2}{c}{Cifar100} & \multicolumn{2}{c}{MiniImagenet}   \\
 & ACC & FR & ACC & FR   \\

\midrule

PVBF-RR  & $21.7\pm0.7$& $11.4\pm0.6$  & $20.9\pm0.4$& $\textbf{9.3}\pm0.5$ \\
PVBF-ZS  & $22.4\pm0.5$& $\textbf{11.0}\pm0.5$  & $20.8\pm0.5$& $10.4\pm0.7$ \\
PVBF-RS  & $\textbf{22.8}\pm0.7$& $11.2\pm0.6$  & $\textbf{21.8}\pm0.5$& $9.8\pm0.5$ \\

\bottomrule
\hline
\end{tabular*}
\caption{Long task sequence experiment results on split Cifar100 and split MiniImagenet using PVBF with different standardization methods. All entries are 95\%-confidence intervals over 15 runs.}
\label{tab:long2}
\end{table}
\section*{Acknowledgement}
This work was supported in part by the National Key Research and Development Project under Grant 2023YFC3806000, in part by the National Natural Science Foundation of China under Grant U23A20382.
\bibliographystyle{elsarticle-num} 
\bibliography{ref.bib}



\end{document}